\documentclass[preprint,5p,times]{elsarticle}
\usepackage{natbib}
\usepackage{amsmath}
\newtheorem{definition}{Definition}
\usepackage{amsfonts,amssymb}
\usepackage{subfigure}
\usepackage{graphicx}
\usepackage{multirow}
\usepackage[normalem]{ulem}
\useunder{\uline}{\ul}{}
\usepackage{amssymb}
\usepackage{bbding}
\usepackage{threeparttable}
\journal{Knowledge-Based Systems}

\begin{document}
\begin{frontmatter}
\title{LSTTN: A Long-Short Term Transformer-based Spatio-temporal Neural Network for Traffic Flow Forecasting}

\author[a,b]{Qinyao~Luo}
\author[a]{Silu~He}
\author[a]{Xing~Han}
\author[a]{Yuhan~Wang}
\author[a,b]{Haifeng~Li\corref{cor}}
\cortext[cor]{Corresponding author: Haifeng Li (lihaifeng@csu.edu.cn) } 
\cortext[cor]{Citation: Q. Luo, S. He, X. Han, Y. Wang, H. Li. LSTTN: A Long-Short Term Transformer-based spatiotemporal neural network for traffic flow forecasting. Knowledge-Based Systems. 2024, 293: 111637} 

\address[a]{School of Geosciences and Info-Physics, Central South University, No.932 South Lushan Road, Changsha, 410083, Hunan, China}
\address[b]{Xiangjiang Laboratory, No. 569, YueLu Avenue, Changsha, 410083, Hunan, China}
\begin{abstract}

Accurate traffic forecasting is a fundamental problem in intelligent transportation systems and learning long-range traffic representations with key information through spatiotemporal graph neural networks (STGNNs) is a basic assumption of current traffic flow prediction models. However, due to structural limitations, existing STGNNs can only utilize short-range traffic flow data; therefore, the models cannot adequately learn the complex trends and periodic features in traffic flow. Besides, it is challenging to extract the key temporal information from the long historical traffic series and obtain a compact representation. To solve the above problems, we propose a novel LSTTN (Long-Short Term Transformer-based Network) framework comprehensively considering the long- and short-term features in historical traffic flow. First, we employ a masked subseries Transformer to infer the content of masked subseries from a small portion of unmasked subseries and their temporal context in a pretraining manner, forcing the model to efficiently learn compressed and contextual subseries temporal representations from long historical series. Then, based on the learned representations, long-term trend is extracted by using stacked 1D dilated convolution layers, and periodic features are extracted by dynamic graph convolution layers. For the difficulties in making time-step level prediction, LSTTN adopts a short-term trend extractor to learn fine-grained short-term temporal features. Finally, LSTTN fuses the long-term trend, periodic features and short-term features to obtain the prediction results. Experiments on four real-world datasets show that in 60-minute-ahead long-term forecasting, the LSTTN model achieves a minimum improvement of 5.63\% and a maximum improvement of 16.78\% over baseline models. The source code is available at https://github.com/GeoX-Lab/LSTTN.
\end{abstract}
\begin{keyword}
Traffic forecasting \sep spatiotemporal modeling \sep long-short term forecasting \sep Transformer \sep Mask Subseries Strategy.
\end{keyword}
\end{frontmatter}

\section{Introduction}
\label{sec:introduction}
With the rapid development of urbanization, intelligent transportation system (ITS) plays an increasingly important role in people’s daily travel life, and the traffic flow forecasting system is also an essential component of ITS. Accurate traffic prediction information can help drivers plan their travel routes in advance to minimize the time delay on the road, and also help related departments to arrange personnel in advance to guide the traffic in congested areas.

The task of traffic flow forecasting is challenging due to its complex spatial dependencies and nonlinear temporal relations. 
Traditional traffic flow prediction methods such as ARIMA \cite{yu2004switching} and VAR \cite{chandra2009predictions} require specific assumptions on traffic data.
However, real-world data are often too complex to satisfy those assumptions, limiting the applicability of those methods to a great extent. More importantly, they cannot capture the complex non-linear temporal dependencies and do not take the spatial dependencies in the traffic network into account.
That generates adverse impact on their prediction accuracies in practical applications. In the era with well-developed technologies of sensing and data processing, vast amounts of traffic data become available to city managers. 
Along with the rapid advancement of deep learning, many methods forecasting traffic flow utilizing deep learning and big data have been proposed and dramatically improved the forecasting results compared to traditional methods. Their performances indicate the great potential of deep learning-based, data-driven traffic flow forecasting methods. 
Such methods have become the mainstream forecasting methods today. 
Early spatiotemporal prediction methods adopted CNNs to model the spatial dependecies and achieved better prediction performance than traditional methods \cite{zhang2017deep,shi2015convolutional}. Since a traffic network is naturally a non-Euclidean graph structure, with the proposal of graph neural networks (GNNs), researchers have started to employ them to model the spatial dependecies in traffic networks, and they have combined GNNs with temporal models such as RNNs and 1-dimensional (1D) CNNs to build traffic flow prediction models \cite{li2017diffusion,yu2017spatio}.

\begin{figure}[htbp]
\centering
\includegraphics[width=8.5 cm]{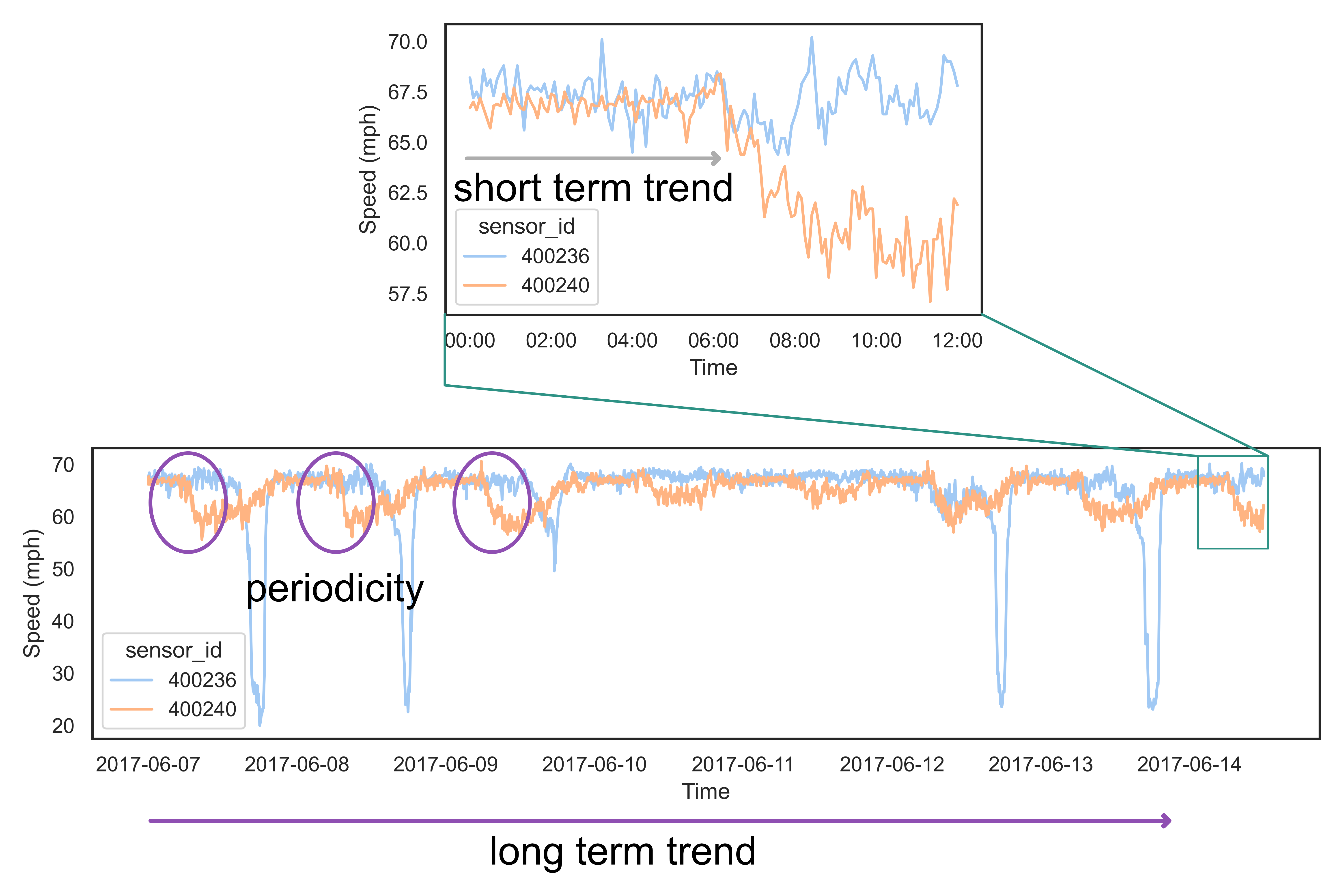}
\caption{A snapshot of a historical time series of sensor 400236 and sensor 400240 in the PEMS-BAY dataset. The sequences in the upper figure correspond to the green window in the lower figure. The short-term historical trend of 00:00-0–6:00 is not sufficient for the model to accurately predict future traffic flow changes, but considering the long-term trend and periodicity in the long historical series can help the model better determine future trends.}
\label{fig:intro}
\end{figure} 

However, the vast majority of these models are limited to using short historical series (e.g., 12 time steps or 1 hour of historical traffic flow features) as model input, but the information in short-term data is often insufficient to adequately represent the rich temporal features present in historical traffic data. Using the upper figure in Figure \ref{fig:intro} as an example, to predict the traffic flow after 06:00, even if we use 6-hour-long historical data from 00:00 to 06:00, it is still difficult for the model to accurately determine the different trends in traffic flow at different locations by relying only on the short-time trend information provided by this part of the historical data since the trends and values of the historical series recorded by the two sensors are also very close. The trend and periodicity features in long historical series can be of great help to the model in determining the complex trends in traffic flow. As shown in the lower figure in Figure \ref{fig:intro}, it is easier for the model to infer different future trends based on the different long-term trend features of the two series and the recurring temporal features in the purple circles. In addition, effectively learning the key temporal information in long historical series is also an issue worth considering. It is inefficient to directly input each time step of long historical series into the model; in addition, since traffic flow series have certain continuity, there is often redundant information in long historical series, and the model should have the ability to efficiently extract the key representations from long series.

To solve the above problems, we introduce long historical series into the traffic flow forecasting model so that the model can make a comprehensive judgment about future traffic flow fluctuations with the help of long-term trend information. We further propose a novel traffic flow forecasting framework called LSTTN (\textbf{L}ong-\textbf{S}hort \textbf{T}erm \textbf{T}ransformer-based \textbf{N}etwork), which can take short-term trends into account while obtaining long-term trends and periodic features from long-term key temporal information-rich representations. The contributions of this paper can be summarized as follows:

\begin{enumerate}
    \item We propose a traffic flow forecasting framework that integrates long- and short-term temporal features and can effectively utilize the long-term trend and periodic features in long historical series, LSTTN.
    \item Specific long-range dependecies-capturing components are designed under the proposed framework. The transformer model is pretrained by a mask reconstruction task to obtain compressed and contextual subseries temporal representations. Then, long-term trend features are extracted from the obtained subseries representations by stacked 1D convolutional layers, and a spatial-based graph convolution is adopted to obtain periodic features in long historical series.
    \item We evaluate the proposed framework on four real-world datasets, and the experimental results show that the LSTTN model outperforms the baseline models for all prediction horizons. The visualization of the prediction results shows that the LSTTN model is robust to missing data and abrupt changes.The source code of this paper is available at https://github.com/GeoX-Lab/LSTTN.
\end{enumerate}

The rest of this paper is organized as follows. Section \ref{sec:related-work} reviews previous traffic flow forecasting methods, Section \ref{sec:methodology} formally describes the traffic flow forecasting task and presents the overall framework and design details, and Section \ref{sec:experiments} evaluates the method through multiple sets of experiments. Finally, Section \ref{sec:conclusion} concludes the paper and provides possible future research directions.

\section{Related Work}
\label{sec:related-work}
Traffic flow prediction is a fundamental task in intelligent transportation systems. According to the modeling approach, mainstream traffic flow prediction methods can be classified as statistical-based methods and data-driven spatial-temporal graph neural network methods.

\subsection{Statistical-based Traffic Flow Forecasting Methods}
Most early traffic flow forecasting methods are based on statistics, such as historical average (HA) \cite{smith1997traffic} models, ARIMA \cite{yu2004switching} models, and Kalman filtering \cite{gao2013application}. The historical weighted average model is based on the temporal periodicity of traffic flow and takes the average value of traffic features at the same time-of-day as the prediction result. ARIMA requires input time series to be stationary after differencing and captures the linear relationship in the differenced time series. The Kalman filter completes the prediction by modeling the state of the dynamic system.

Most of the early traffic flow forecasting models migrated the methods that were applied to ordinary time series, ignoring the complex interactions between roads in the traffic network and lacking consideration of spatial dependencies in traffic flow data.

\subsection{STGNN-based Traffic Flow Forecasting Methods}

\subsubsection{Structural Design of STGNN-based Traffic Flow Forecasting Methods}

In recent years, the use of neural networks to model the spatiotemporal dependecies in traffic flow and train the models on massive data has become a new trend in the field of traffic flow prediction, and has achieved better results than previous works \cite{jiang2022graph}. Some methods apply convolutional neural networks (CNNs) to model the spatial relationships of traffic data organized in the form of grids \cite{ma2017learning,liu2017short}. They consider the Euclidean distance and spatial location relationships between roads, but lack a measure of topological relationships. However, road networks can be naturally expressed in the form of graphs, which are more suitable for describing the topological relationships between roads, and the distance between roads can be represented as the weights of the edges. The emergence of graph neural networks (GNNs) has solved the problem that previous deep learning methods had difficulty in terms of modeling non-Euclidean data structures, and GNNs have been applied to various different tasks ever since \cite{kipf2016variational,zhang2018end,li2022curvature}.

GNNs can be classified into two categories, namely spectral-based GNNs and spatial-based GNNs \cite{wu2020comprehensive}. Bruna et al. first proposed the spectral CNN \cite{bruna2013spectral}, which generalizes CNNs to non-Euclidean space, but the process of spectral decomposition of Laplacian matrices suffers from excessive computational complexity. ChebNet \cite{defferrard2016convolutional} applies Chebyshev polynomials to approximate complex spectral convolutions, effectively reducing the complexity of the spectral convolution and reducing the number of parameters, thus it is less likely to overfit the data. GCN \cite{kipf2016semi} can be regarded as a first-order ChebNet with regularization. It is computationally efficient and easy to stack in multiple layers. Spatial-based GNNs take the perspective of local spatial associations between nodes, and describe the process of graph convolution as an aggregation of information from the central node and its neighboring nodes. Glimer et al. summarized the operations (e.g. aggregation, readout, etc.) of the spatial-based GNNs and generalized a class of message passing neural networks (MPNN) \cite{gilmer2017neural}. The GCN can also be regarded as a spatial-based message passing neural network, where the weight of the messages from neighbor nodes during aggregation is fixed and determined by the degree of the nodes. GAT \cite{velivckovic2017graph} introduces a multihead self-attention mechanism into graph neural networks to dynamically assign the weight of messages from neighboring nodes during aggregation. Thus, compared with GCN, GAT can better focus on the information from important nodes.

On the basis of GNNs, studies further introduced the modeling of temporal dependecies in traffic flow and proposed numerous spatiotemporal graph neural network models for traffic forecasting tasks \cite{bai2021a3t,he2022stgc,cai2020traffic,park2020st,roy2021unified,zhao2019t}. Depending on the modeling of the temporal dependecies, we classify STGNNs into RNN-based, CNN-based and Transformer-based STGNNs.

(1) RNN-based STGNNs. DCRNN \cite{li2017diffusion} uses a gated recurrent unit (GRU), a variant of the RNN, to extract temporal features from data, as the GRU achieves similar performance to LSTM with fewer parameters. A diffusion convolution neural network (DCNN) is adopted to simulate the effect of the spatiotemporal changes that spread from one node to other nodes in a traffic network. T-GCN \cite{zhao2019t} instead combines GRU with GCN, and the results in two common scenarios, namely, on highways and urban roads, show that the performance of T-GCN exceeds baseline models. AST-GCN \cite{zhu2021ast} and KST-GCN \cite{zhu2022kst} are further proposed to incorporate external factors. However, due to the limitations of the linear chain structure of RNNs, the parameters of RNN-based STGNNs cannot be trained and updated in parallel, so the backpropagation process of these models is very time-consuming \cite{yu2017spatio}. On the other hand, as the features from earlier steps may be forgotten, RNN-based models have difficulty in capturing long-range temporal dependencies in data.

(2) CNN-based STGNNs. Some works consider 1D convolutions to model the temporal dependecies. STGCN \cite{yu2017spatio} adopts a 1D causal convolution and gated linear unit (GLU) to extract temporal features from traffic data, and combines them with ChebNet to build a spatiotemporal feature extractor. Graph WaveNet \cite{wu2019graph} constructs an adaptive adjacency matrix that captures hidden spatial dependencies in traffic data through an end-to-end manner without any prior knowledge and utilizes 1D dilated causal convolution to enlarge the model’s temporal receptive field. ASTGCN \cite{guo2019attention} introduces a spatiotemporal attention mechanism to capture long-range temporal dependencies and the implied spatial connections in the road network and combines it with spatial graph convolution and temporal 1D convolutions to model the spatiotemporal dynamics in traffic data. Overall, CNN-based models no longer suffer from long training time and vanishing gradient problems, and 1D convolutions help CNN-based models learn temporal dependencies from traffic data.

(3) Transformer-based STGNNs. Due to structural constraints, the direct input of long sequences into the models mentioned above may cause the problem of excessive computational complexity, so the above methods tend to use short-range traffic flow data (e.g., one-hour historical traffic flow information) to predict future traffic flows. However, the inherent periodicity and the long-term trend information in traffic flows are difficult to extract directly from short historical time series. Recently, with the emergence of Transformer \cite{vaswani2017attention}, some research has begun to focus on how to use it to fully mine long-range dependencies in historical time series based. While ensuring that the model parameters can be trained in parallel, Transformer enables direct connection between the individual time steps of the input sequence independent of the linear chain structure of RNNs or the receptive field in CNNs. This approach can help Transformer-based models better capture long-range dependencies in traffic data. Traffic Transformer \cite{cai2020traffic} designs various encoding strategies for the positional encoding mechanism to help Transformer learn day-level and week-level periodicity from traffic flow data. STTN \cite{xu2020spatial} incorporates the graph convolution process into spatial Transformer, and together with temporal Transformer, STTN can capture long-range dependencies from traffic data. Although the above methods take the dependecies between distant time steps into account, they still only use short historical time series to make predictions. Other works consider improving the original Transformer model to make predictions based on long sequences. For example, Informer \cite{zhou2021informer} proposed ProbSparse attention to reduce the complexity of the original self-attention mechanism. In addition, a generative decoder is employed to avoid the error accumulation problem that arises with increasing time steps. However, these works ignore the spatial dependecies present in traffic networks and capture mostly trend features while failing to better capture the inherent periodicity in traffic data \cite{zeng2022transformers}.

\subsubsection{Learning Paradigm of STGNN-based Traffic Flow Forecasting Methods}

In general, STGNNs mostly adopt a supervised learning paradigm and use only traffic flow features of future moments as supervision signals to guide the model’s training process. In recent years, the rapid development of self-supervised learning, which learns representations from massive data without additional labels, has led to a large number of applications in computer vision \cite{he2022masked,chen2021self}, natural language processing \cite{devlin2018bert,radford2019language}, and graph representation learning \cite{li2023augmentation,zhu2022high,zhu2022alleviating}. Additionally, STGNNs can also benefit from additional supervision signals. Considering the relatively small size of most of the publicly available traffic datasets, STGCL \cite{liu2022contrastive} designs various data augmentation methods and uses contrastive learning as an auxiliary task to guide STGNNs to better distinguish different spatiotemporal features in historical data. ST-SSL \cite{ji2022spatio}, on the other hand, uses the perspective of spatiotemporal heterogeneity and augments the graph structure and traffic features. STEP \cite{shao2022pre} passes the long-range representations obtained from Transformer into a graph structure learning module to learn spatial dependencies from long historical sequences but does not sufficiently account for the long-term trend and the periodic features that are unique to long sequences.

Given the above background, in this paper, we consider exploiting the long-term trend and periodicity from long historical traffic data and expect the model to produce accurate prediction results based on the long-range and compact representations learned from the long-term sequences.

\section{Methodology}
\label{sec:methodology}

\subsection{Problem Definition}

\begin{definition}[Traffic network]
    A traffic network $G$ can be expressed as $G = (V,E,A)$. The node set $V$ ($\left|V\right| = N$) denotes sensors or road crossings in the traffic network, and the edge set $E$ can be obtained according to the spatial distances of sensors or the topological relations of roads.
\end{definition}

\begin{definition}[Traffic forecasting problem]
    The traffic forecasting problem can be expressed as predicting the traffic features $F$ future time steps ahead given the historical features of $H$ time steps and the adjacency matrix $\mathbf{A}$:
    \begin{equation}
        \resizebox{.9\hsize}{!}{
            $
            \left[\mathbf{X}_{t}, \mathbf{X}_{t+1}, \dots, \mathbf{X}_{t+F-1}\right] = f\left(\mathbf{A}, \left[\mathbf{X}_{t-H}, \mathbf{X}_{t-H+1}, \dots, \mathbf{X}_{t-1}\right]\right)
            $
        }
    \end{equation}
    where $\mathbf{X} \in \mathbb{R}^{N \times T}$ denotes the traffic feature matrix, and $T$ denotes the total length of the series in the dataset. The vast majority of current traffic forecasting methods $f$ accept a short historical series, e.g., $H = 12$. In this paper, we attempt to utilize the temporal patterns of traffic flow from a longer historical series, which can be expressed as follows:
    \begin{equation}
        \left[\mathbf{X}_{t}, \mathbf{X}_{t+1}, \dots, \mathbf{X}_{t+F-1}\right] = f\left(\mathbf{A}, \mathbf{X}_{\mathrm{short}},\mathbf{X}_{\mathrm{long}}\right)
    \end{equation}
    where $\mathbf{X}_{\mathrm{short}} = \left[\mathbf{X}_{t-S}, \mathbf{X}_{t-S+1}, \dots, \mathbf{X}_{t-1}\right]$, $S$ denotes the length of the short-term historical series given to the model. $\mathbf{X}_{\mathrm{long}} = \left[\mathbf{X}_{t-L}, \mathbf{X}_{t-L+1}, \dots, \mathbf{X}_{t-1}\right]$. $L$ denotes the length of the long historical series.
\end{definition}

\subsection{LSTTN Framework}

\begin{figure}[htbp]
\centering
\includegraphics[width=8.5 cm]{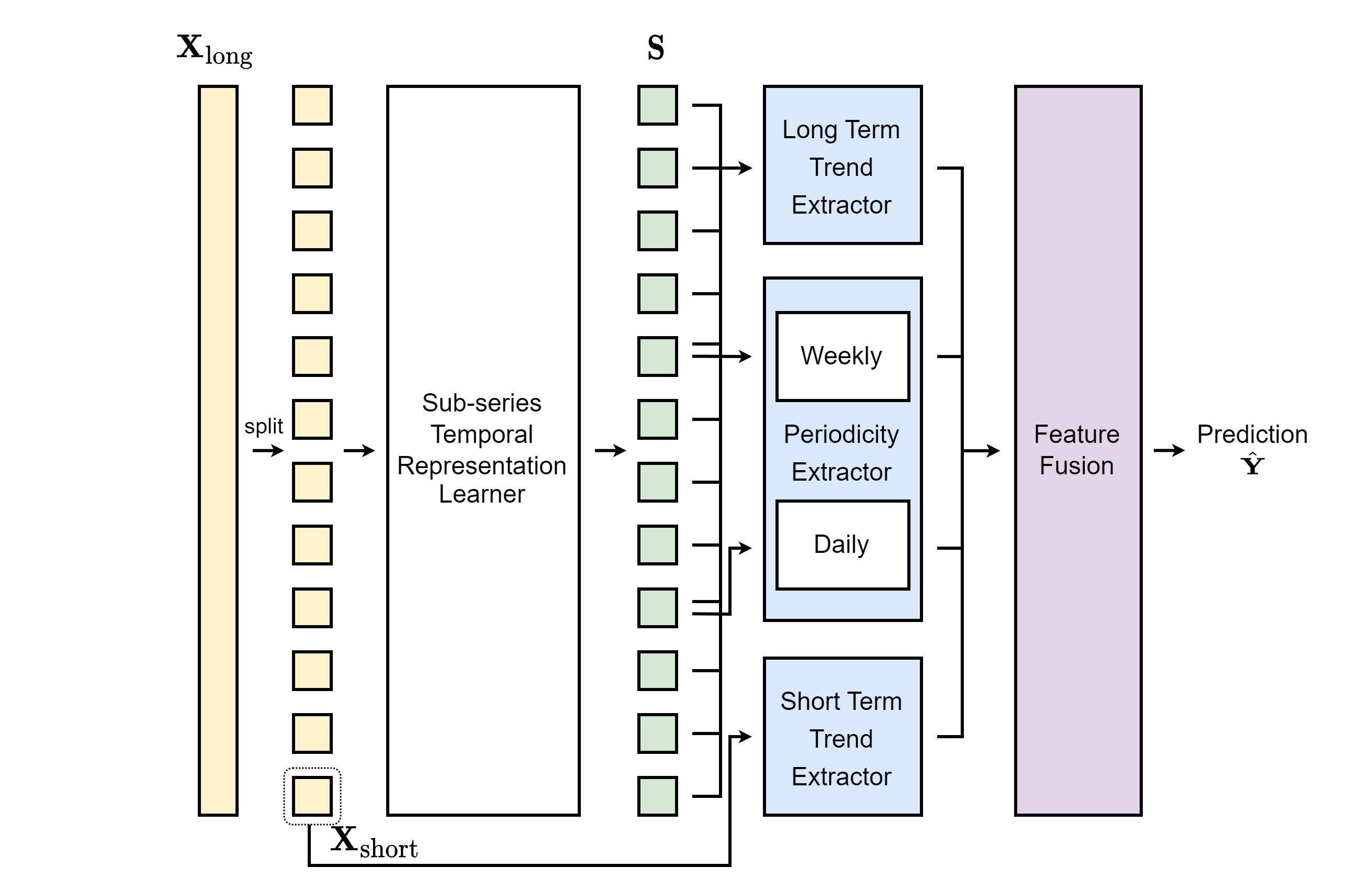}
\caption{An overview of the LSTTN framework. The long historical time series $\mathbf{X}_\mathrm{long}$ is split into subseries of equal lengths, and the last subseries is taken as the short-term historical time series $\mathbf{X}_\mathrm{short}$. The subseries-level temporal representation $\mathbf{S}$, which is rich in key temporal information, is learned by the subseries temporal representation learner, and then, long-term trend and periodic features are extracted from $\mathbf{S}$. Meanwhile, time-step level short-term trend are extracted directly from $\mathbf{X}_\mathrm{short}$. Finally, the long-term and short-term features are fused to obtain the prediction results.}
\label{fig:overview}
\end{figure}

The overall LSTTN framework is shown in Figure \ref{fig:overview}. We adopt a subseries temporal representation learner to extract subseries-level temporal representations from long-term series and design a long-term trend extractor, periodicity extractor and short-term trend extractor to obtain long- and short-term temporal features from subseries-level representations. Specifically:
\begin{enumerate}
    \item Subseries temporal representation learner (STRL). As it is inefficient to learn time-step level fine-grained representations for long-term series, we instead learn subseries-level representations from the temporal context by using a mask reconstruction task.
    \item Long-term trend extractor. This module expands the model’s temporal receptive field by stacking dilated 1-dimensional convolutional layers to capture long-term trends from subseries representations.
    \item Periodicity extractor. This module captures the inherent periodicity in traffic flow from the representations of the same period of time from the previous same week-of-day and the previous day.
    \item Short-term trend extractor. As the subseries representations are coarse-grained, we cannot obtain time-step level forecasting results from those representations. The module directly utilizes the existing STGNN model to learn local fine-grained trend features from the short-term historical series.
\end{enumerate}

\subsection{Masked Sub-series Transformer}

The fundamental purpose of the masked subseries Transformer (MST) is to infer the contents of masked subseries from a small number of subseries and their temporal context so that the model can efficiently learn compressed, context information-rich subseries representations from long time series. The design of MST consists of two basic problems: (1) the masking strategy and (2) the model for learning representations.

(1) Masking strategy

Two important elements of the masking strategy of the MST need to be considered: the basic unit for masking and the mask ratio.

(a) The basic unit for masking. Existing methods usually use a 5-minute time step as the basic unit of input data, which is inefficient for capturing the trend of long time series. Inspired by \cite{shao2022pre,li2023ti,nie2022time}, we divide the long series into equal-length subseries containing multiple time steps, and the subseries are taken as the basic unit of the model input.

(b) Mask ratio. Both BERT \cite{devlin2018bert} and MAE \cite{he2022masked} learn the essential semantic information in the data by mask reconstruction. In natural language, the semantic information of words is richer, and sentences rarely have redundant parts and have high information density, so BERT sets a lower masking ratio of 15\%. The information density of image data is relatively low, and there is spatial continuity of pixel points, so even if MAE masks off 75\% of the pixels, the main content of the image can still be inferred. Long-term traffic flow data are similar to images with temporal continuity and low information density, so a relatively high masking ratio is needed. We apply 75\% random masking in this paper.

(2) Model for learning representations

Another issue to consider is which model to choose to learn the subseries representations. For time series, one of the differences between Transformer and temporal models such as RNN and 1D CNN is that the inputs to each time step in Transformer are directly connected to each other. 
Regardless of increases in time step, Transformer learns representations with previous temporal features being considered.
In this paper, the Transformer encoder is adopted as the STRL.

\begin{figure}[htbp]
\centering
\includegraphics[width=8.5 cm]{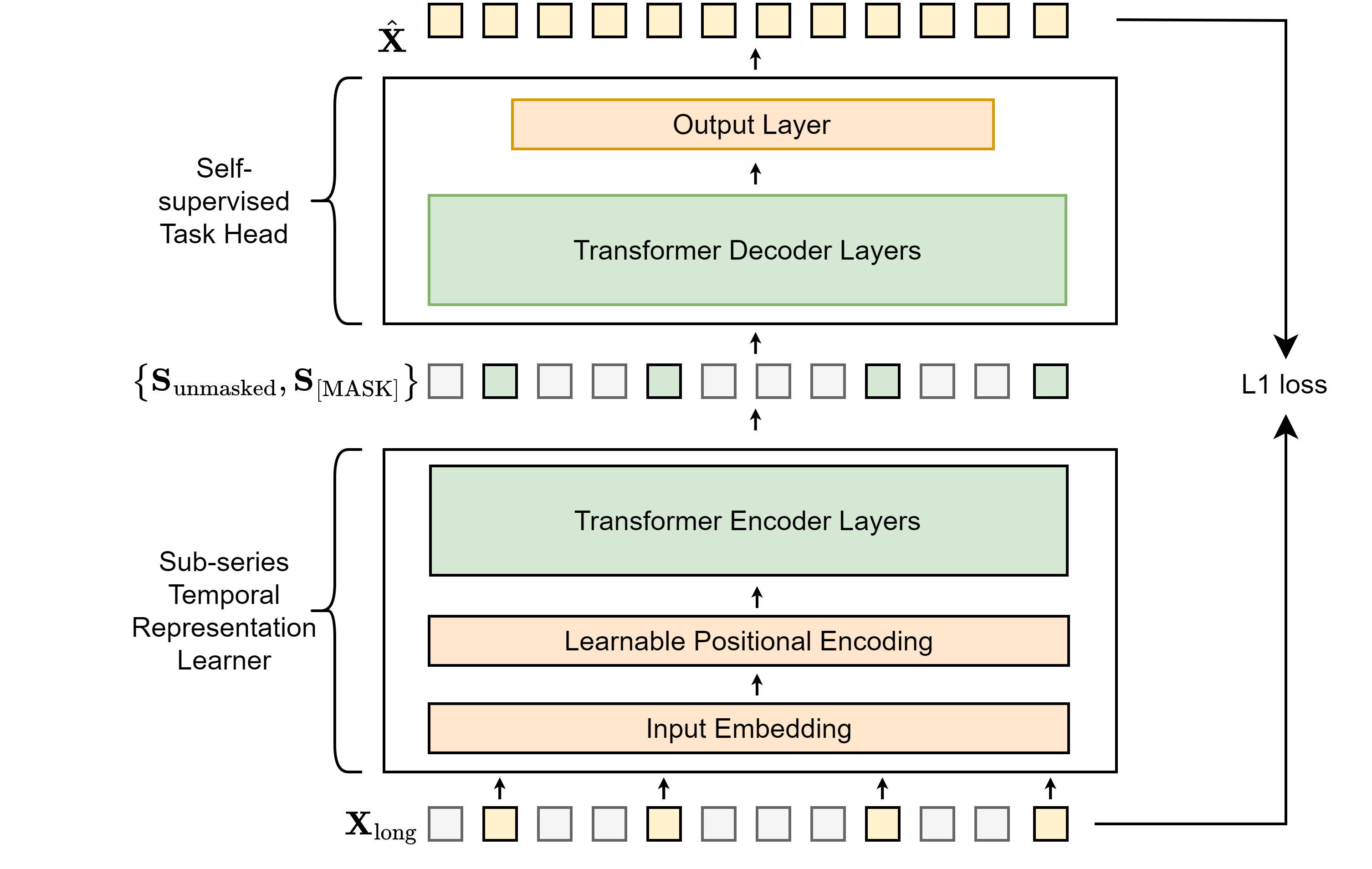}
\caption{The structure of masked subseries Transformer}
\label{fig:mst}
\end{figure}   

As shown in Figure \ref{fig:mst}, MST consists of two parts: the STRL and the self-supervised task head. The STRL learns the subseries temporal representation, and the self-supervised task head reconstructs the complete long series based on the temporal representation of the unmasked subseries and the mask tokens. Specifically, first, $\mathbf{X}_\mathrm{long}=\left[\mathbf{X}_{t-L}, \mathbf{X}_{t-L+1}, \dots, \mathbf{X}_{t-1}\right]$, the long historical series is divided into nonoverlapping subseries containing $S$ time steps, so the number of subseries is $N=L/S$. We randomly mask 75\% of the subseries, and the set of masked subseries is denoted as $\mathbf{X}_\mathrm{masked}$. The remaining data are denoted as $\mathbf{X}_\mathrm{unmasked}$ and are used as the input to STRL:
\begin{equation}
    \mathbf{S}_\mathrm{unmasked} = \mathrm{STRL}(\mathbf{X}_\mathrm{unmasked})
\end{equation}
where $\mathbf{S}_\mathrm{unmasked}$ denotes the representations of $\mathbf{X}_\mathrm{unmasked}$, which is the output by STRL. The self-supervised task head consists of a Transformer layer and a linear output layer and can reconstruct the complete long series given $\mathbf{S}_\mathrm{unmasked}$ and the learnable mask token $\mathbf{S}_{\mathrm{[MASK]}}$ as follows:

\begin{equation}
    \resizebox{.9\hsize}{!}{
        $\left[\hat{\mathbf{X}}_{\mathrm{masked}}, \hat{\mathbf{X}}_{\mathrm{unmasked}}\right] = \mathrm{TaskHead}\left(\left[\mathbf{S}_{\mathrm{unmasked}}, \mathbf{S}_{\mathrm{[MASK]}}\right]\right)$
    }
\end{equation}

The goal of the pretraining stage is to make the error between the reconstructed $\hat{\mathbf{X}}_{\mathrm{masked}}$ and the masked truth values as small as possible, so only the masked subseries are considered when calculating the loss:
\begin{equation}
    \resizebox{.9\hsize}{!}{
        $
        \mathcal{L}_{\mathrm{pretrain}}\left(\hat{\mathbf{X}}_{\mathrm{masked}}, \mathbf{X}_{\mathrm{masked}}; \Theta_{\mathrm{T}}\right) = \left\Vert \hat{\mathbf{X}}_{\mathrm{masked}} - \mathbf{X}_{\mathrm{masked}} \right\Vert
        $
    }
\end{equation}
where $\Theta_{\mathrm{T}}$ represents the learnable parameters of the whole Transformer.

\subsection{Long-term Trend Extractor}

The relatively small amount of information in a short-term historical series is not sufficient for the model to infer complex future changes in traffic flow, while the long historical series can help the model determine the fluctuation of traffic flow in future moments. For this reason, we design a long-term trend extractor to extract the long-term trend features in traffic flow from the subseries temporal representations.

\begin{figure}[htbp]
\centering
\includegraphics[width=7 cm]{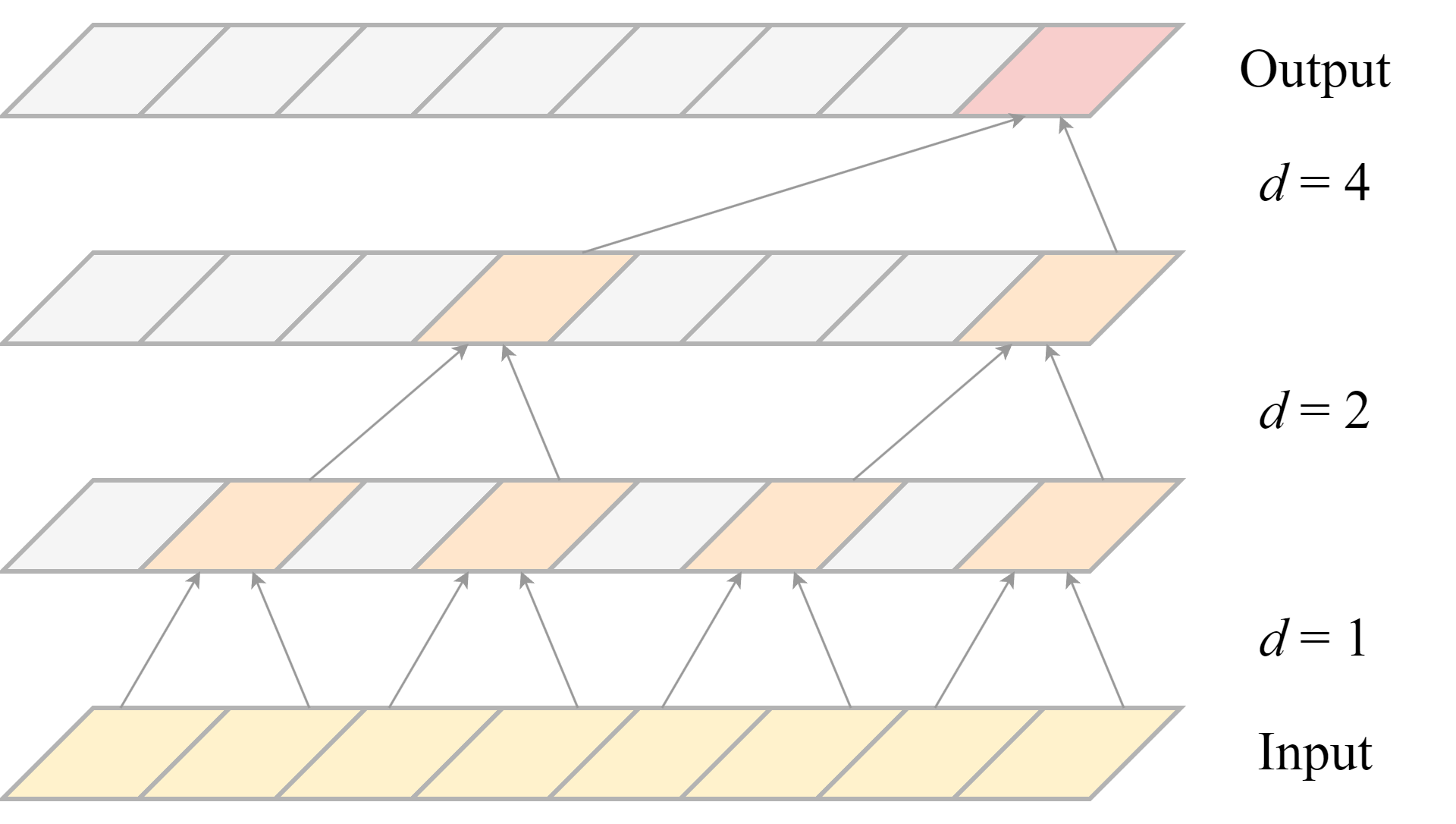}
\caption{A demonstration of the 1-dimensional dilated convolution layer. The yellow parallelograms at the bottom represent input data, and the red parallelogram represents the output. Layers in the middle are hidden layers. As the number of layers increases, the receptive field grows exponentially, enabling the efficient capture of long-range features.}
\label{fig:dilated-conv}
\end{figure}   

Commonly adopted basic structures for temporal feature extraction include the RNN and 1-dimensional CNN. However, as RNNs cannot process the features of each time step in parallel and are prone to the problem of gradient vanishing and gradient explosion, RNNs have difficulty handling long series. Ordinary 1-dimensional CNNs have a limited receptive field, and increasing that field requires stacking multiple CNN layers, which leads to a large increase in the number of model parameters with increasing model depth. As shown in Figure \ref{fig:dilated-conv}, we adopt stacked 1-dimensional dilated convolution layers \cite{yu2015multi} as the long-term trend extractor. The receptive field of the module grows exponentially as the number of 1-dimensional dilated convolution layers increases, which can capture trend features efficiently while avoiding problems such as gradient vanishing. The dilated convolution operation can be expressed as follows:
\begin{equation}
    \mathbf{x} *_d \mathbf{C}(m) = \sum^{k - 1}_{j=0}\mathbf{C}(m)\mathbf{x}(m - d \times j)
\end{equation}
where $m$ denotes the $m$-th element in series $\mathbf{x}$, $\mathbf{C} \in \mathbb{R}^k$ represents the convolution kernel, and d represents the dilation rate. In this paper, a convolution layer can be expressed as follows:
\begin{equation}
    \mathbf{D}_i = \mathrm{Conv}_{i}(\mathbf{D}_{i - 1}) = \mathrm{MaxPool}(\mathrm{gelu}(\mathbf{D}_{i - 1} *_d \mathbf{C}_i))
\end{equation}
where the max pooling operation is adopted to reduce dimensions. The dilation rate $d$ of the $i$-th layer is set to $2^i$. When $i = 1$, the input to the module is the set of subseries temporal representations S, i.e., $\mathbf{D}_1 = \mathbf{S} = \left[\mathbf{S}_1, \mathbf{S}_2, \dots, \mathbf{S}_N \right]$. The output of the last convolution layer is taken as the long-term trend feature $\mathbf{H}_{\mathrm{long}}$.

\subsection{Periodicity Extractor}

\begin{figure}[htbp]
\centering
\includegraphics[width=8.5 cm]{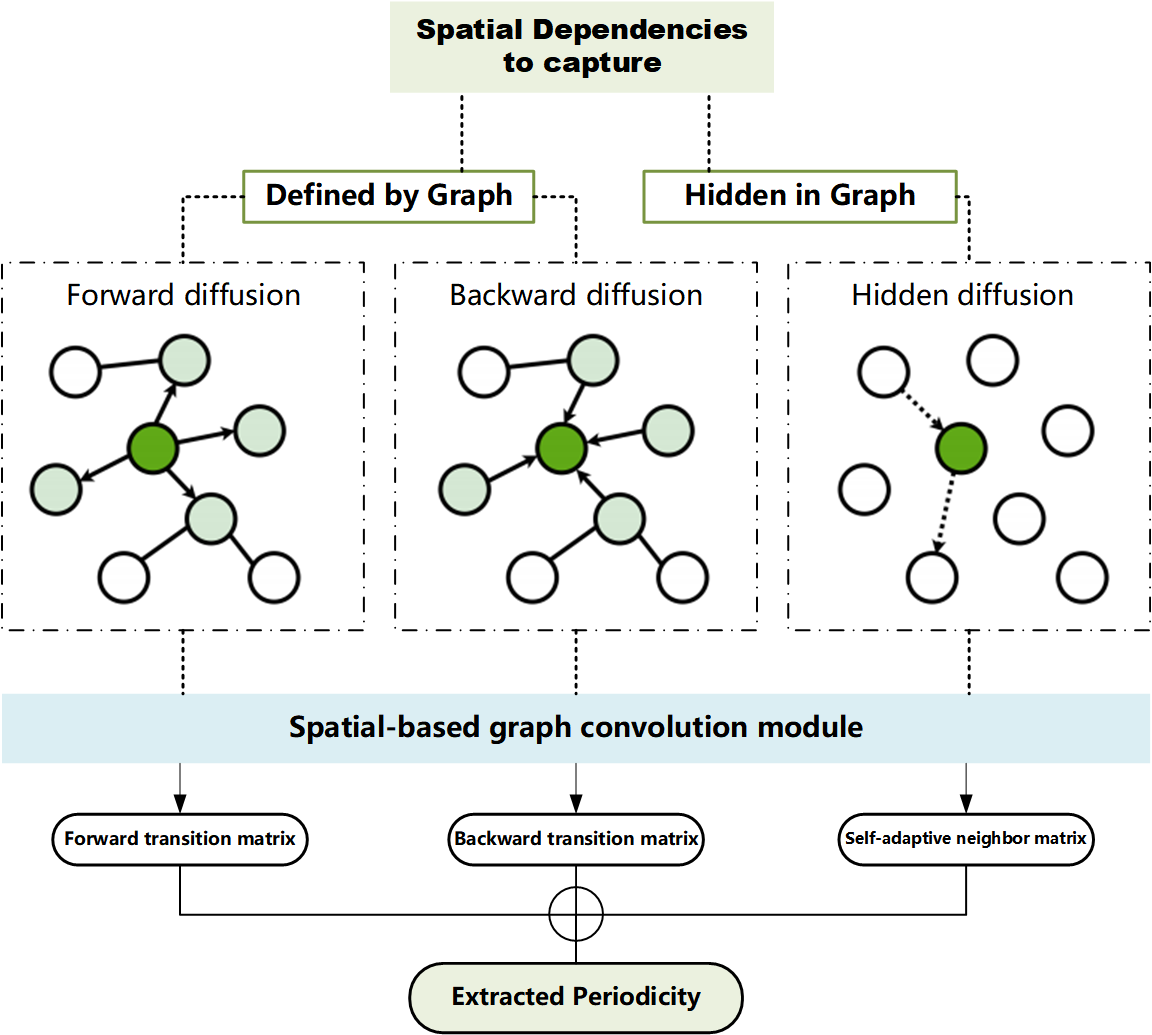}
\caption{A demonstration of the spatial-based graph convolution module. In this module, the spatial dependencies between two nodes are supposed to come from three diffusions, forward diffusion and backward diffusion that are defined by graph structure, as well as the diffusion hidden in graph. If there is no available graph structure, this module will only capture the dependencies which are from hidden diffusion and depicted by the self-adaptive neighbor matrix.}
\label{fig:dynamic-graph-conv}
\end{figure}

Periodicity is also an important pattern of traffic flow. Generally, there is daily and weekly periodicity in traffic flow, and traffic in the same time period from different dates can show very similar spatiotemporal patterns. In this paper, we build a periodicity extractor to capture periodic spatiotemporal features from the subseries representations of the corresponding time period in the previous week and the previous day. As the representations of the subseries already contain temporal information and no longer have temporal dimensions, this module mainly considers extracting the spatial dependencies among the input features across different time steps of each node while retaining the temporal information of the subseries. 
Let the duration of a day corresponds to a series of $l$ segments; then, the representations of the corresponding moments from the previous week and the previous day can be denoted as $\mathbf{S}_\mathrm{week} = \mathbf{S}_{N-7 \times l}$ and $\mathbf{S}_\mathrm{day} = \mathbf{S}_{N-l}$, respectively ($N$ denotes the number of subseries which equals $L/S$). As shown in Figure \ref{fig:dynamic-graph-conv}, $\mathbf{S}_\mathrm{week}$ and $\mathbf{S}_\mathrm{day}$ are passed into a spatial-based graph convolution module for the obtainment of periodic temporal features $\mathbf{H}_\mathrm{day}$ and $\mathbf{H}_\mathrm{day}$. This module is similar to the one proposed in \cite{wu2019graph}. As Figure \ref{fig:dynamic-graph-conv} indicates, the graph convolution module combines the spatial dependencies defined by the structure of the traffic graph and the spatial dependencies hidden in the graph.
\begin{equation}
    \resizebox{.9\hsize}{!}{
        $
        \mathbf{H}_\mathrm{week} = \sum^{K}_{k=0}\mathbf{P}^k_f\mathbf{S}_\mathrm{week}\mathbf{W}_{k1} + \mathbf{P}^k_b\mathbf{S}_\mathrm{week}\mathbf{W}_{k2} + \mathbf{A}^k_\mathrm{adp}\mathbf{S}_\mathrm{week}\mathbf{W}_{k3}
        $
    }
\end{equation}
\begin{equation}
    \resizebox{.9\hsize}{!}{
        $
        \mathbf{H}_\mathrm{day} = \sum^{K}_{k=0}\mathbf{P}^k_f\mathbf{S}_\mathrm{day}\mathbf{W}_{k4} + \mathbf{P}^k_b\mathbf{S}_\mathrm{day}\mathbf{W}_{k5} + \mathbf{A}^k_\mathrm{adp}\mathbf{S}_\mathrm{day}\mathbf{W}_{k6}
        $
    }
\end{equation}
where $\mathbf{P}_f = \frac{\mathbf{A}}{\sum^n_{j=1}\mathbf{A}_{ij}}$ and $\mathbf{P}_b = \frac{\mathbf{A}^\mathrm{T}}{\sum^n_{j=1}\mathbf{A}^\mathrm{T}_{ij}}$ correspond to the forward diffusion and backward diffusion of graph signals, respectively, and represent corresponding transition matrices. The power of the matrices $k$ represents the number of steps in the diffusion process. $\mathbf{W}$ denotes the weight matrices. $\mathbf{A}_\mathrm{adp}$ is a self-adaptive neighbor matrix which can be considered as the transition matrix of a hidden diffusion process. $\mathbf{A}_\mathrm{adp} = \mathrm{Softmax}\left(\mathrm{ReLU}\left(\mathbf{E}_1\mathbf{E}^\mathrm{T}_2\right)\right)$. $\mathbf{E}_1$ and $\mathbf{E}_2$ represent the source node embedding and target node embedding, respectively. The spatial dependency weight between the source node and target node is derived by multiplying $\mathbf{E}_1$ and $\mathbf{E}_2$. $\mathrm{ReLU()}$ is for eliminating weak dependencies, and $\mathrm{Softmax()}$ is for normalization.

\subsection{Short-term Trend Extractor}

The granularity of the subseries-level temporal representations extracted from long series is coarse, which makes it difficult for the model to accurately predict the fine-grained traffic flow trends at the time-step level. In addition, there is a strong temporal correlation between the future short-term traffic flow and the historical short-term traffic flow, so the short-term trend needs to be modeled separately.

It has been widely proved that STGNNs excel at capturing fine-grained features from short-term series \cite{wu2019graph,li2017diffusion, yu2017spatio}. Basically, they complete the feature capture in two steps. First, they learn spatial features and temporal features separately via spatial learning networks and temporal learning networks. Then they fuse two kinds of features through a certain spatiotemporal fusion neural architecture. 
Given the acknowledged capability of STGNNs, we can directly adopt an existing STGNN such as Graph WaveNet \cite{wu2019graph}, DCRNN \cite{li2017diffusion} or STGCN \cite{yu2017spatio}, as the short-term trend extractor to obtain finer-grained short-term trend $\mathbf{H}_\mathrm{short}$:

\begin{equation}
    \mathbf{H}_\mathrm{short} = \mathrm{STGNN}\left(\mathbf{A}, \mathbf{X}_\mathrm{short}\right)
\end{equation}

where $\mathbf{X}_\mathrm{short}$ represents the last subseries in the long series, i.e., the subseries closest to the current moment, $\mathbf{A}$ represents the neighbor matrix, STGNN() represents the adopted STGNN model.

\subsection{Feature Fusion}

To comprehensively consider the long- and short-term features in the long historical series, the previously obtained long-term trend features, periodic features and short-term trend features are fused to obtain the final prediction result $\hat{\mathbf{Y}}$:
\begin{equation}
    \hat{\mathbf{Y}} = \mathrm{MLP}\left(\mathrm{MLP}\left(\mathbf{H}_\mathrm{long} \Vert \mathbf{H}_\mathrm{week} \Vert \mathbf{H}_\mathrm{day}\right) \Vert \mathbf{H}_\mathrm{short}\right)
\end{equation}
where $\Vert$ denotes the concatenation operation. The aim of the traffic flow forecasting task is to make the model output as close as possible to the true values, so L1 loss is chosen as the objective function. It is expressed as follows:
\begin{equation}
    \mathcal{L}_\mathrm{forecast}\left(\hat{\mathbf{Y}}, \mathbf{Y}; \Theta\right) = \left\Vert \hat{\mathbf{Y}} - \mathbf{Y} \right\Vert
\end{equation}
where $\mathbf{Y} = \left[ \mathbf{X}_t, \mathbf{X}_{t+1}, \dots, \mathbf{X}_{t+F-1} \right]$ represents the ground truth and $\Theta$ denotes the set of learnable parameters in the whole model.

\section{Experiments}
\label{sec:experiments}

\subsection{Datasets}

Four public real-world traffic flow datasets were chosen to fully evaluate the performance of our model.

(1) METR-LA. This dataset contains traffic speed data collected from 207 sensors on the Los Angeles County freeway system, ranging from March 1, 2012 to June 30, 2012.

(2) PEMS-BAY. This dataset consists of traffic flow speeds collected from 325 sensors on highways in bay area, ranging from Jan 1, 2017 to May 31, 2017.

(3) PEMS04. This dataset consists of traffic flow, occupancy and speed data, collected from 307 sensors on 29 highways in California. The time span of the dataset is from January to February 2018.

(4) PEMS08. This dataset contains traffic data collected from 170 sensors on 8 highways in California. Similar to PEMS04, it consists of three features: flow, occupancy and speed. The time span of the dataset is from July to August 2018.

Detailed statistical information about the four datasets is provided in Table \ref{dataset}. To provide a fair comparison, the dataset division in this paper is kept consistent with other examples in the literature. For METR-LA and PEMS-BAY, the ratio of the training, validation and test sets is 7:2:1, and for PEMS04 and PEMS08, it is 6:2:2.
\begin{table*}[htbp]
\caption{\centering {Stastistics of datasets}}
\label{dataset}
\centering
\begin{tabular}{cccccc}
\hline
                & Dataset   & Nodes    & Edges     & Time interval   & Samples   \\ \hline
                & METR-LA     & 207     & 1722    & 5mins            & 34272  \\
                & PEMS-BAY   & 325   & 2694   & 5mins   & 52116   \\
                & PEMS04   & 307   & 987   & 5mins   & 16992    \\
                & PEMS08   & 170   & 718   & 5mins   & 17856   \\ \hline
\end{tabular}
\end{table*}

\subsection{Experimental Setup}

\subsubsection{Evaluation Metrics}

Let the number of samples in the dataset be $n$. In addition, let $\mathbf{Y}_i = \mathbf{X}_{t:t+F} = \left[\mathbf{X}_t, \mathbf{X}_{t+1}, \dots, \mathbf{X}_{t+F-1}\right]$ denote the ground truth of the $i$-th sample in the dataset and $\hat{\mathbf{Y}}_i$ denote the prediction result provided by the model. We adopt the following metrics to evaluate the performance of our model:

\begin{itemize}
    \item Root mean squared error (RMSE):
    \begin{equation}
        \mathrm{RMSE} = \sqrt{
            \frac{1}{n}
            \sum^n_{i=1}
            \left(
            \hat{\mathbf{Y}}_i -
            \mathbf{Y}_i
            \right)^2
        }
    \end{equation}
    \item Mean absolute error (MAE):
    \begin{equation}
        \mathrm{MAE} = \frac{1}{n}\sum^n_{i=1} \left| \hat{\mathbf{Y}}_i - \mathbf{Y}_i \right|
    \end{equation}
    \item Mean absolute percentage error (MAPE):
    \begin{equation}
        \mathrm{MAPE} = \frac{1}{n}\sum^n_{i=1} \left| \frac{\hat{\mathbf{Y}}_i - \mathbf{Y}_i}{\mathbf{Y}_i} \right|
    \end{equation}
\end{itemize}

For consistency with previous works and to provide a fair comparison with other baselines, we omit missing data when calculating the metrics above.

\subsubsection{Baseline Models}

We compare our method with the following nine traffic forecasting models. The former five are widely adopted baseline models and the latter four are models proposed in nearly two years. A brief description of each model is provided as below:

(1) DCRNN \cite{li2017diffusion}, a combination of diffusion convolutional neural network and a gated recurrent unit;

(2) STGCN \cite{yu2017spatio}, a spatiotemporal network that stacks 1D gated convolution blocks and graph convolution blocks;

(3) Graph WaveNet \cite{wu2019graph}, a spatiotemporal network that utilizes adaptive dependency matrix and merges 1D causal convolution and graph convolution to capture spatiotemporal dependencies;

(4) GMAN \cite{zheng2020gman}, a multihead attention network that captures the spatial relationship between different nodes and the temporal relationship between different time steps;

(5) MTGNN \cite{wu2020connecting}, a model capable of learning multi-scale spatio-temporal features without prior knowledge about graph structure via a graph learning layer;

(6) DDSTGCN \cite{sun2022dual}, a dual dynamic STGCN that utilizes edge features to transform traffic flow graphs into hypergraphs;

(7) STID \cite{shao2022spatial}, a model that encodes information based on simple multi-layer perceptrons and makes predictions through a regression layer;

(8) DSTET \cite{sun2023transformer}, a transformer network which enhances the characterzation of spatial-temporal features through decoupling the spatial and temporal embedding;

(9) DAGN \cite{ouyang2023domain}, a domain adversarial graph neural network which captures node-pair adjacent relationships to enable dynamic aggregation of spatial-temporal information.
\subsubsection{Experimental Environment and Hyperparameter Settings}

All of the experiments are conducted on the Linux platform equipped with an Intel Xeon Gold 5218R and a NVIDIA RTX A6000 GPU, and the models are implemented with PyTorch \cite{paszke2019pytorch}. The length of the input sequence is set to two weeks, which is equivalent to 4032 time steps for datasets with a 5-minute interval between each data point. The batch size is set to 32, and the number of training iterations was 100. We use the Adam optimizer \cite{kingma2014adam} to train the model, and a multistep learning rate scheduler is adopted to control how the learning rate decreases as the training process progresses. The hidden dimension of the long-term trend extractor and periodicity extractor is set to 4. The hyperparameters of the STGNN are set according to those provided in the original paper, and the STGNN used in the experimental part of this paper is Graph WaveNet.

\subsection{Experimental Results}

\subsubsection{Prediction Accuracy}

We compare the LSTTN model with the nine baseline models on the four real-world traffic datasets. The results are shown in Table \ref{table:accuracy}. To provide fair comparisons, for the five widely adopted baseline models, we adopt the experimental results from the LibCity \cite{wang2021libcity} benchmark. In the benchmark, the hyperparameters of those baseline models have been carefully adjusted, and the results are better than those given in the original papers in most cases. For the four baseline models proposed in recent two years, the experimental results are extracted from \cite{sun2023transformer} and \cite{ouyang2023domain}. The former paper provides the top results of DDSTGCN, STID and DSTET under the same experimental setting, the latter presents the best performance of DAGN.

\begin{table*}[htbp]
\centering
\caption{\centering {Prediction performances of traffic forecasting models on multiple datasets. The best results are marked in bold, and the second-best results are underlined.}}
\label{table:accuracy}
\resizebox{\linewidth}{!}{
\begin{tabular}{cccccccccccccc}
\hline
\multicolumn{2}{c}{\multirow{2}{*}{\textbf{}}}                                 & \multicolumn{3}{c}{METR-LA}                         & \multicolumn{3}{c}{PEMS-BAY}                        & \multicolumn{3}{c}{PEMS04}                             & \multicolumn{3}{c}{PEMS08}                          \\
\multicolumn{2}{c}{}                                                           & 15 min          & 30 min          & 60 min          & 15 min          & 30 min          & 60 min          & 15 min           & 30 min           & 60 min           & 15 min          & 30 min          & 60 min          \\ \hline
\multirow{3}{*}{DCRNN}                                                  & RMSE & 4.61            & 5.33            & 6.25            & 2.18            & 2.84            & 3.58            & 29.50            & 31.35            & 34.20            & 22.08           & 23.64           & 25.99           \\
                                                                        & MAE  & 2.49            & 2.74            & 3.08            & 1.10            & 1.33            & 1.59            & 18.51            & 19.70            & 21.56            & 14.30           & 15.22           & 16.70           \\
                                                                          & MAPE & 6.19\%          & 7.12\%          & 8.44\%         & 2.21\%          & 2.82\%          & 3.58\%          & 12.60\%          & 13.45\%          & 14.87\%          & 9.31\%          & 9.86\%         & 10.79\%         \\ \hline
\multirow{3}{*}{STGCN}                                                  & RMSE & 4.80            & 5.51            & 6.34            & 2.46            & 2.98            & 3.57            & 29.53            & 30.49            & 31.86            & 22.87           & 23.88           & 25.32           \\
                                                                        & MAE  & 2.59            & 2.82            & 3.11            & 1.32            & 1.50            & 1.73            & 18.70            & 19.28            & 20.11            & 14.80           & 15.33           & 16.22           \\
                                                                        & MAPE & 6.61\%          & 7.43\%          & 8.53\%         & 2.77\%          & 3.23\%          & 3.86\%          & 13.06\%          & 13.42\%          & 13.98\%          & 9.81\%          & 10.11\%         & 10.64\%         \\ \hline
\multirow{3}{*}{Graph WaveNet}                                          & RMSE & 4.84            & 5.54            & 6.33            & 2.20            & {\ul{2.83}}            & {\ul{3.50}}            & {\ul{27.70}}      & \textbf{28.60}      & {\ul{29.81}}      & {\ul{20.70}}     & {\ul{21.76}}     & \textbf{23.34}           \\
                                                                        & MAE  & 2.58            & 2.83            & 3.14            & 1.10            & 1.32            &  1.57      & 17.20      & 17.76      & 18.56      & {\ul{13.03}}     & {\ul{13.55}}     & {\ul{14.34}}     \\
                                                                        & MAPE & 6.58\%          & 7.58\%          & 8.89\%         & 2.23\%          & 2.82\%          & {\ul{3.51\%}}          & 12.02\%    & 12.49\%    & 13.35\%    & 8.76\%    & {\ul{8.98\%}}    & {\ul{9.76\%}}         \\ \hline
\multirow{3}{*}{GMAN}                                                   & RMSE & 4.99            & 5.51            & 6.17      & 2.46            & 2.99            & 3.56            & 28.91            & 29.66            & 30.77            & 22.40           & 23.23           & 24.61           \\
                                                                        & MAE  & 2.80            & 3.01            & 3.31            & 1.33            & 1.54            & 1.78            & 18.47            & 18.88            & 19.59            & 14.92           & 15.29           & 16.11           \\
                                                                        & MAPE & 7.44\%          & 8.21\%          & 9.26\%         & 2.73\%          & 3.26\%          & 3.90\%          & 15.66\%          & 15.78\%          & 16.12\%          & 12.31\%         & 12.52\%         & 13.15\%         \\ \hline
\multirow{3}{*}{MTGNN}                                                  & RMSE & 4.55      & {\ul{5.19}}      & {\ul{6.01}}            & 2.21            & 2.85            & 3.55            & 27.97            & 28.92            & 30.23            & 21.05           & 22.10           & 23.58           \\
                                                                        & MAE  & 2.46      & {\ul{2.69}}      & {\ul{3.00}}      & 1.11            & 1.34            & 1.59            & 17.41            & 17.97            & 18.73            & 13.55           & 14.08           & 14.96           \\
                                                                        & MAPE & 6.87\%    & 7.00\%    & {\ul{8.14\%}}    & 2.25\%          & 2.86\%          & 3.58\%          & 12.75\%          & 13.31\%          & 13.81\%          & 10.00\%          & 9.99\%         & 10.56\%         \\ \hline
\multirow{3}{*}{DDSTGCN}                                                & RMSE & 5.01      & 6.02      & 7.13          & 2.71            & 3.64            & 4.37            & 28.38            & 29.54            & 30.69            & 21.14           & 22.23           & 24.12           \\
                                                                        & MAE  & 2.64      & 3.00      & 3.44      & 1.29            & 1.60            & 1.89            & 17.69            & 18.51            & 19.51            & 13.54           & 14.02           & 14.70           \\
                                                                        & MAPE & 6.76\%    & 8.12\%    & 9.74\%    & 2.69\%          & 3.62\%          & 4.46\%          & 12.34\%          & 12.70\%         & 13.46\%          & 8.82\%          & 9.71\%         & 10.41\%        \\ \hline
\multirow{3}{*}{STID}                                                & RMSE & 5.53      & 6.60      & 7.54          & 2.81            & 3.72            & 4.40            & 28.48            & 29.86            & 31.79            & 21.66           & 23.57           & 25.89           \\
                                                                        & MAE  & 2.80      & 3.18      & 3.55      & 1.30            & 1.62            & 1.89            & 17.51            & 18.29            & 19.58            & 13.28           & 14.21           & 15.58           \\
                                                                        & MAPE & 7.70\%    & 9.40\%    & 10.95\%    & 2.73\%          & 3.68\%          & 4.47\%          & 12.00\%          & 12.46\%         & 13.38\%          & {\ul{8.62\%}}         & 9.24\%         & 10.33\%        \\ \hline
\multirow{3}{*}{DSTET}                                                & RMSE & {\ul{4.52}}      & 5.22      & 6.08          & \textbf{2.17}            & 2.85            & 3.60           & \textbf{27.38}            & {\ul{28.66}}            & \textbf{29.77}            & \textbf{20.24}           & \textbf{21.63}           & 23.53           \\
                                                                        & MAE  & {\ul{2.46}}      & 2.71      & 3.03      & \textbf{1.06}           & {\ul{1.29}}            & {\ul{1.57}}            & \textbf{16.90}            & {\ul{17.68}}            & {\ul{18.48}}            & \textbf{12.25}           & \textbf{13.21}           & \textbf{14.11}           \\
                                                                        & MAPE & {\ul{6.05\%}}    & {\ul{6.95\%}}    & 8.30\%    & {\ul{2.13\%}}          & {\ul{2.74\%}}          & 3.54\%          & {\ul{11.59\%}}          & {\ul{12.44\%}}         & {\ul{12.71\%}}          & \textbf{8.03\%}          & \textbf{8.83\%}         & \textbf{9.30\%}        \\ \hline
\multirow{3}{*}{DAGN}                                                 & RMSE & 5.03            & 6.00            & 7.49            & 2.58            & 3.22            & 4.21            & 29.68            & 32.02            & 38.87            & 23.56           & 25.46           & 31.97           \\
                                                                      & MAE  & 2.67      & 3.03      & 3.67      & 1.28            & 1.49            & 1.86            & 19.13            & 20.53            & 23.63            & 15.16           & 16.68           & 20.94           \\
                                                                        & MAPE & 6.85\%    & 8.16\%    & 10.55\%    & 2.75\%          & 3.38\%          & 4.50\%          & 13.85\%          & 14.31\%          & 15.55\%          & 10.87\%          & 11.29\%         & 14.30\%         \\ \hline
\multirow{3}{*}{\begin{tabular}[c]{@{}c@{}}LSTTN\\ (Ours)\end{tabular}} & RMSE & \textbf{4.42}   & \textbf{5.10}   & \textbf{5.92}   & {\ul{2.18}}   & \textbf{2.76}   & \textbf{3.41}   & 27.74   & 28.67   & 30.00   & 20.78  & 21.89  & {\ul{23.47}}  \\ 
                                                                        & MAE  & \textbf{2.42}   & \textbf{2.65}   & \textbf{2.96}   & \textbf{1.06}   & \textbf{1.26}   & \textbf{1.48}   & {\ul{17.06}}   & \textbf{17.62}   & \textbf{18.46}   & 13.17  & 13.71  & 14.54  \\ 
                                                                        & MAPE & \textbf{5.94\%} & \textbf{6.80\%} & \textbf{7.93\%} & \textbf{2.11\%} & \textbf{2.63\%} & \textbf{3.24\%} & \textbf{11.34\%} & \textbf{11.68\%} & \textbf{12.17\%} & 8.63\% & 9.09\% & 9.77\% \\ \hline
\end{tabular}
}
\end{table*}

According to Table \ref{table:accuracy}, the LSTTN model acquires very high prediction accuracy on all four datasets. Though the performance of the LSTTN model on PEMS08 is slightly inferior to Graph WaveNet and DSTET, on the other three datasets, the advantage of the LSTTN model over baseline models is quite significant. Especially, on METR-LA, the dataset with the largest number of samples, the LSTTN model outperforms all nine baseline models in all metrics and for all prediction horizons, On the second largest dataset, PEMS-BAY, in MAPE, even compared to the most comparative baseline model, DSTET, the LSTTN model achieves improvements of 0.94\%, 4.01\% and 8.47\% at 15 min, 30 min and 60 min horizons, respectively. It is also worth nothing that, on the three datasets, as the prediction horizon increases, the improvements achieved by the LSTTN model in all three metrics become larger. Specifically, when predicting the next 60 min, the LSTTN model achieves improvements of 9.55\% in RMSE, 8.73\% in MAE and 12.92\% in MAPE on METR-LA. Similarly, on PEMS-BAY, improvements of 9.86\% in RMSE, 13.38\% in MAE and 16.78\% in MAPE are observed. And on PEMS04, improvements of 5.63\% in RMSE, 7.05\% in MAE and 13.44\% in MAPE are achieved by the LSTTN model. Those results indicate that the LSTTN model has stronger long-term prediction capability in comparison.

Another thing worth noting is that except the LSTTN model, DSTET is the only one predicting traffic flow with consideration into periodicity. The result that DSTET displays the best performance among nine baseline models corroborates the importance of utilizing periodicity. Since one major difference between DSTET and the LSTTN model is that the LSTTN model also captures long-term trend, the advantage of the LSTTN model over DSTET may be attributed to its introduction of long-term trend for prediction.

\subsubsection{Ablation Studies}

To demonstrate that each module in the LSTTN model can effectively improve the prediction performance, ablation experiments are conducted on the METR-LA dataset. the experimental results are shown in Table \ref{tab:ablation}, where w/o LT represents the original the LSTTN model without a long-term trend extractor, w/o P represents the original model without a periodicity extractor, w/o ST indicates that the short-term trend extractor, i.e., STGNN, was removed from the original model, and ST only indicates that only STGNN was used to generate prediction results.

\begin{table*}[htbp]
\label{tab:ablation}
\caption{Ablation studies}
\centering
\begin{tabular}{cccccccccc}
\hline
\multicolumn{1}{l}{} & \multicolumn{3}{c}{15 min} & \multicolumn{3}{c}{30 min} & \multicolumn{3}{c}{60 min} \\
                     & RMSE   & MAE    & MAPE     & RMSE   & MAE    & MAPE     & RMSE   & MAE    & MAPE     \\ \hline
Ours                 & 4.42   & 2.42   & 5.98\%   & 5.10   & 2.65   & 6.80\%   & 5.92   & 2.96   & 7.93\%   \\
w/o LT               & 4.71   & 2.50   & 6.43\%   & 5.59   & 2.81   & 7.76\%   & 6.74   & 3.23   & 9.57\%   \\
w/o P                & 4.46   & 2.43   & 6.02\%   & 5.21   & 2.69   & 6.98\%   & 6.15   & 3.04   & 8.30\%   \\
w/o ST               & 8.71   & 4.07   & 12.03\%  & 8.72   & 4.07   & 12.04\%  & 8.75   & 4.09   & 12.11\%  \\
ST only              & 4.84   & 2.58   & 6.58\%   & 5.54   & 2.83   & 7.58\%   & 6.33   & 3.14   & 8.89\%  \\ \hline
\end{tabular}
\end{table*}

As seen from the table, (1) the LSTTN model without the long-term trend module has worse performance than the original the LSTTN model for all prediction horizons, and as the horizon increases, the difference in their performances also increases, with a maximum difference of approximately 17\% (60 min, MAPE metric). This result may have been found because the fluctuation in traffic flow is relatively small in a short horizon and has a high correlation with short-term historical features. However, as the horizon becomes longer, the short-term historical features become increasingly insufficient in terms of adequately judging complex future temporal trends, and the long-term trend tends to play a more important role. (2) The performance of the LSTTN model without the periodicity module is weaker than that of the original model for all horizons, which indicates that explicitly modeling the periodicity in traffic flow helps improve the model’s performance. However, the contribution of periodicity to the performance improvement is relatively small compared to that of the long-term trend. (3) The prediction performance of the LSTTN model without the short-term trend module is significantly worse than that of the original model, which illustrates the necessity of retaining the short-term trend module. It is worth noting that the metric values of the LSTTN model without the short-term trend module are very close at different forecast lengths, which may happen because the model that only utilizes coarse subseries representations cannot infer the fine-grained time-step level fluctuations.

\subsubsection{Hyperparameter Analysis}

\begin{figure}[htbp]
\centering  
\subfigure[]{
\label{fig:pretrain-losses-line}
\includegraphics[width=0.3\textwidth]{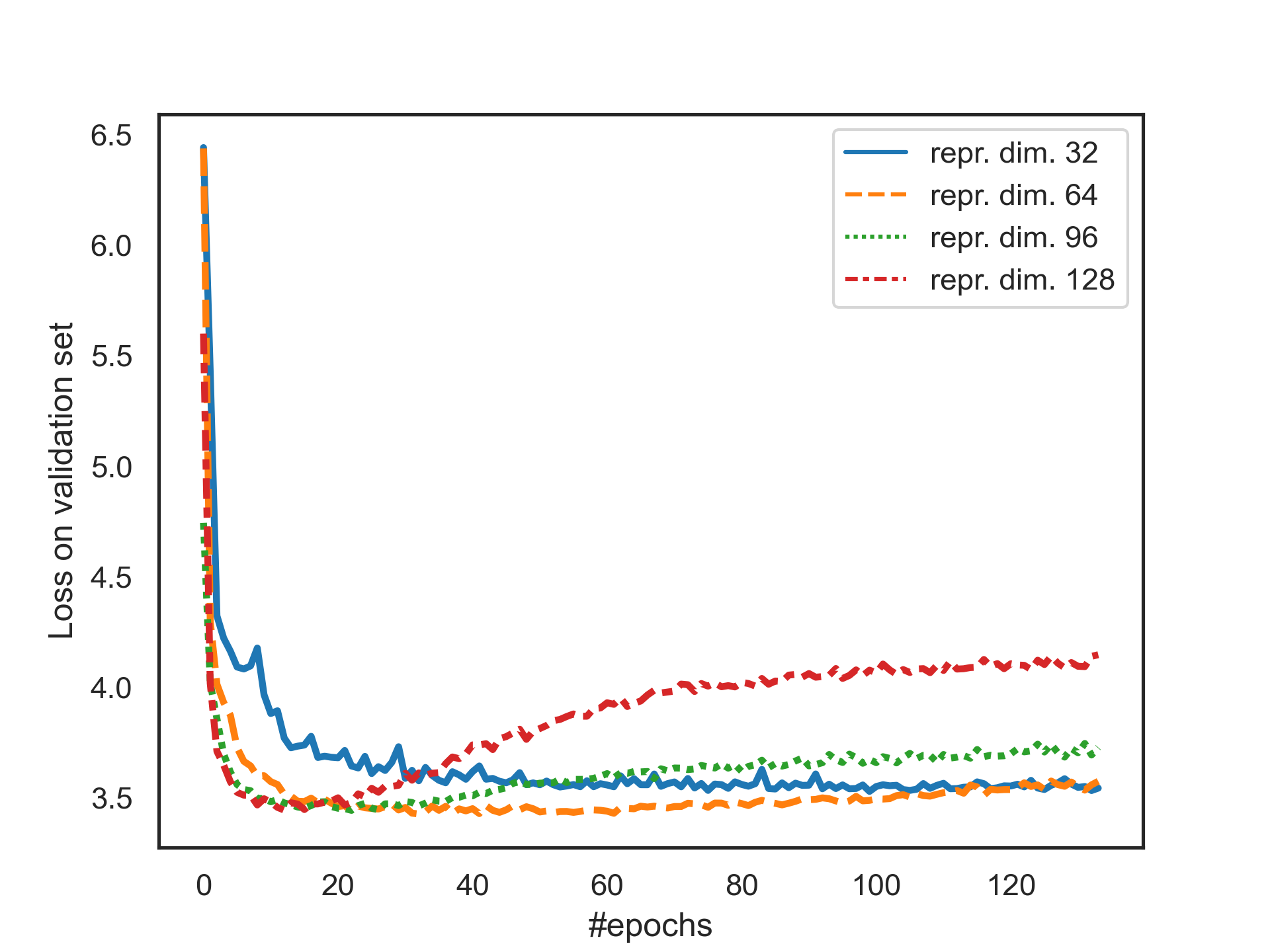}
}
\subfigure[]{
\label{fig:pretrain-losses-bar}
\includegraphics[width=0.15\textwidth]{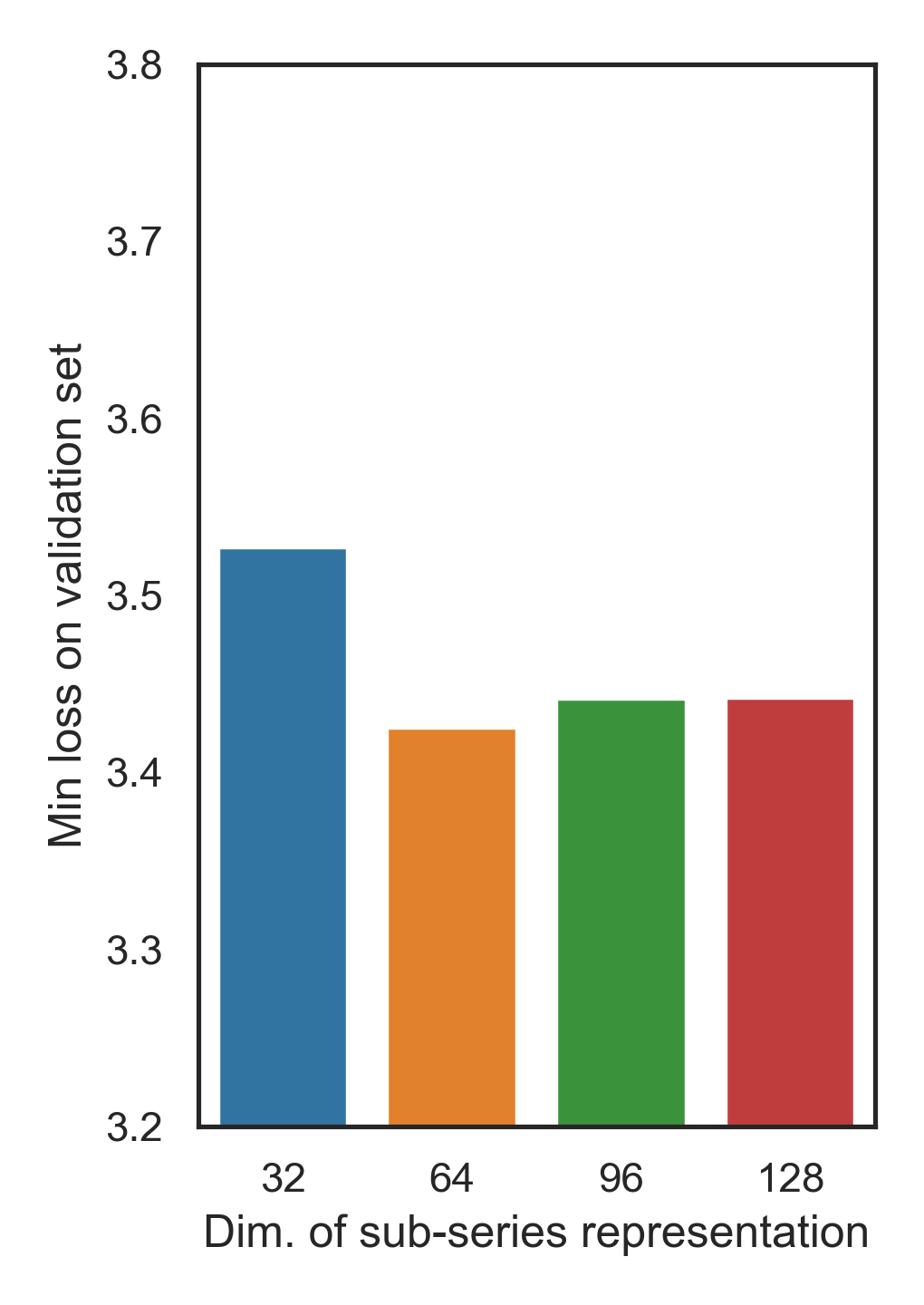}
}
\caption{Loss values of the Transformer model on the validation set during the pretraining stage. (a) Change in validation losses as iteration continues, and (b) the lowest validation loss values during iterations with different dimensions of subseries representations.}
\end{figure}

(1) Dimension of subseries representations. The dimension of each subseries representation output by Transformer, i.e., the dimension of the hidden Transformer layer during the pretraining stage, is an important hyperparameter of the model. We selected the best dimension from [32, 64, 96, 128] and conducted experiments on the METR-LA dataset. To account for model generalizability while avoiding test set information leakage, the validation loss was used as an evaluation metric. A lower loss value indicates a better ability to reconstruct masked subseries from given contexts and to capture the contextual associations in long series. As shown in Figure \ref{fig:pretrain-losses-bar}, with the increase in dimension, the validation loss shows a trend of decreasing and then increasing, and when the dimension is set to 32, the subseries representations cannot adequately express the contextual information and their own temporal information in the long-term sequence. When the dimension is set to 96 or 128, as shown in Figure \ref{fig:pretrain-losses-line}, it becomes easy for the model to overfit the training set, and the loss value is slightly worse than that when the dimension is set to 64. Therefore, we set the dimension of the subseries representations to 64 in this paper.

\begin{figure}[htbp]
\centering  
\subfigure{
\label{fig:hidden-dim-rmse}
\includegraphics[width=0.4\textwidth]{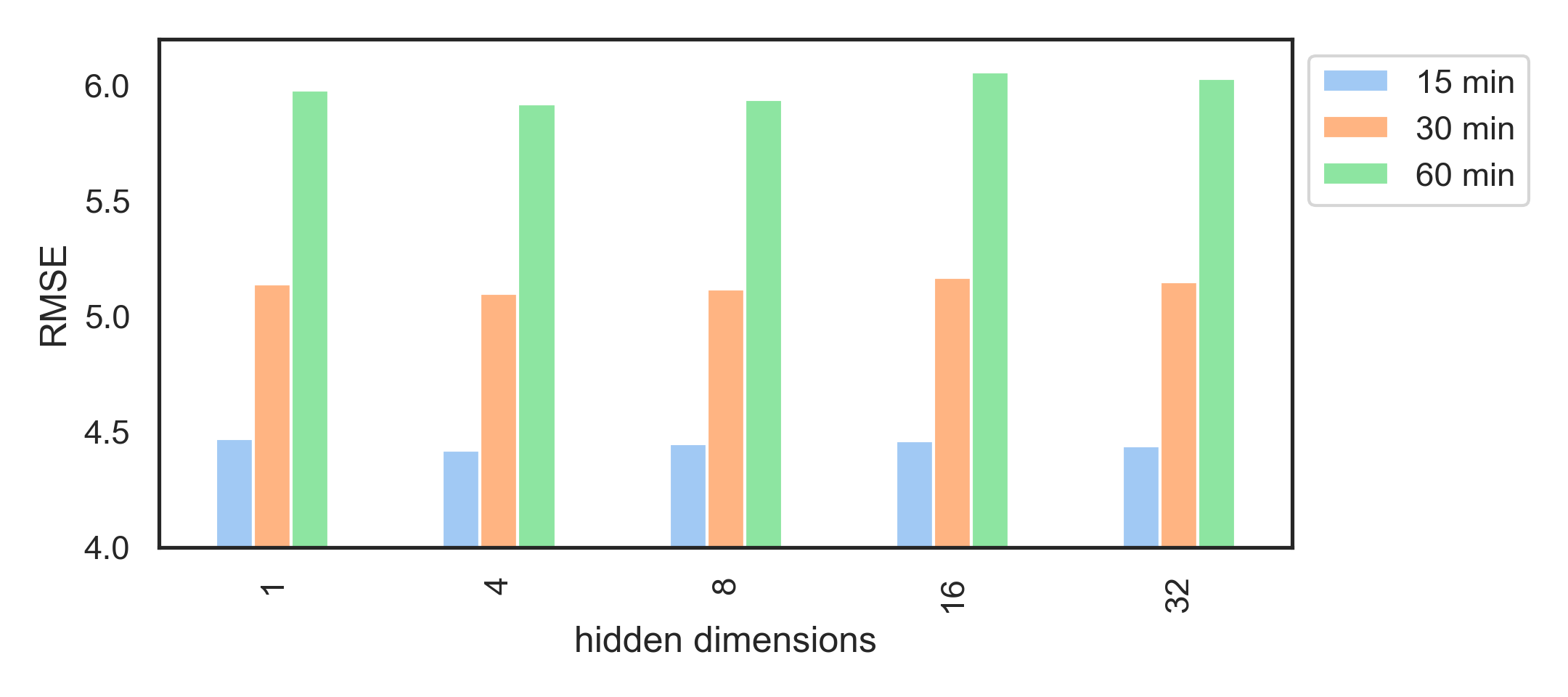}
}
\subfigure{
\label{fig:hidden-dim-mae}
\includegraphics[width=0.4\textwidth]{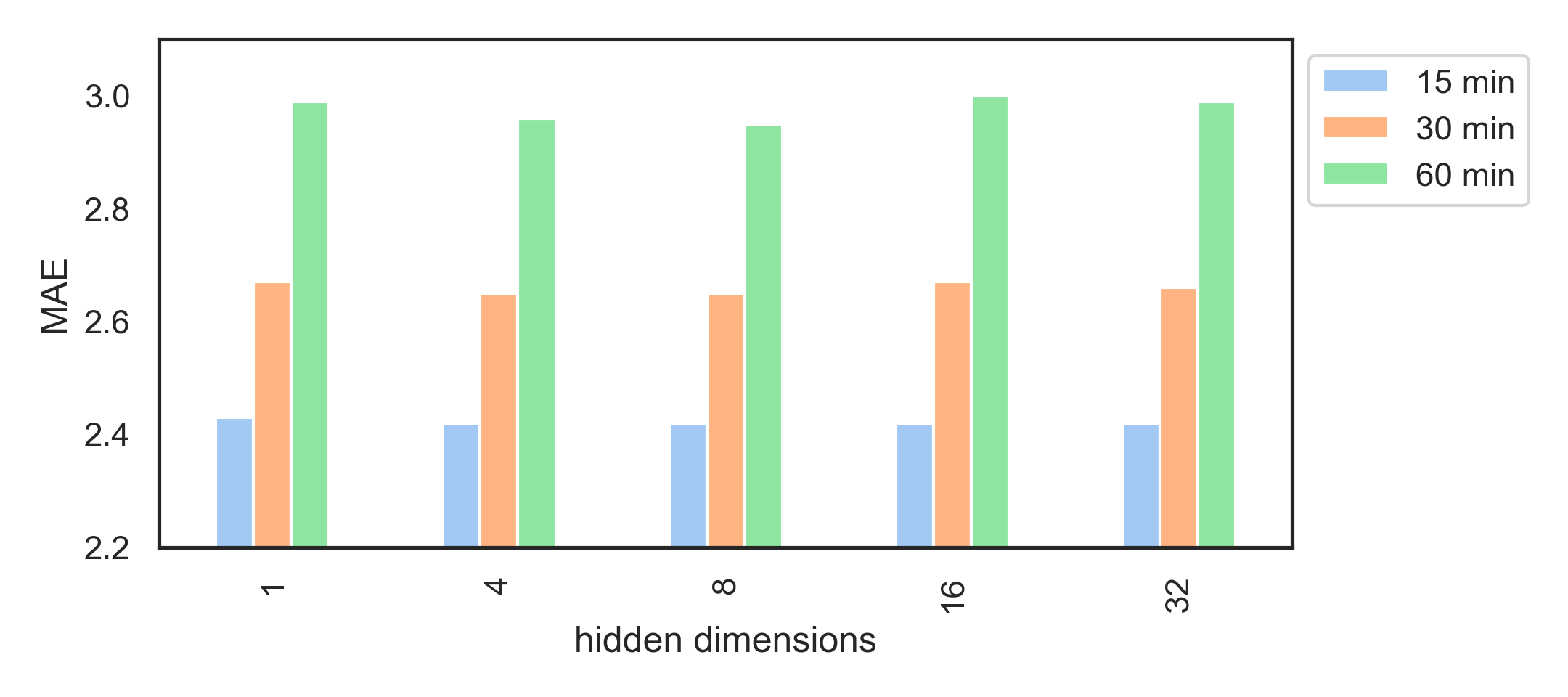}
}
\subfigure{
\label{fig:hidden-dim-mape}
\includegraphics[width=0.4\textwidth]{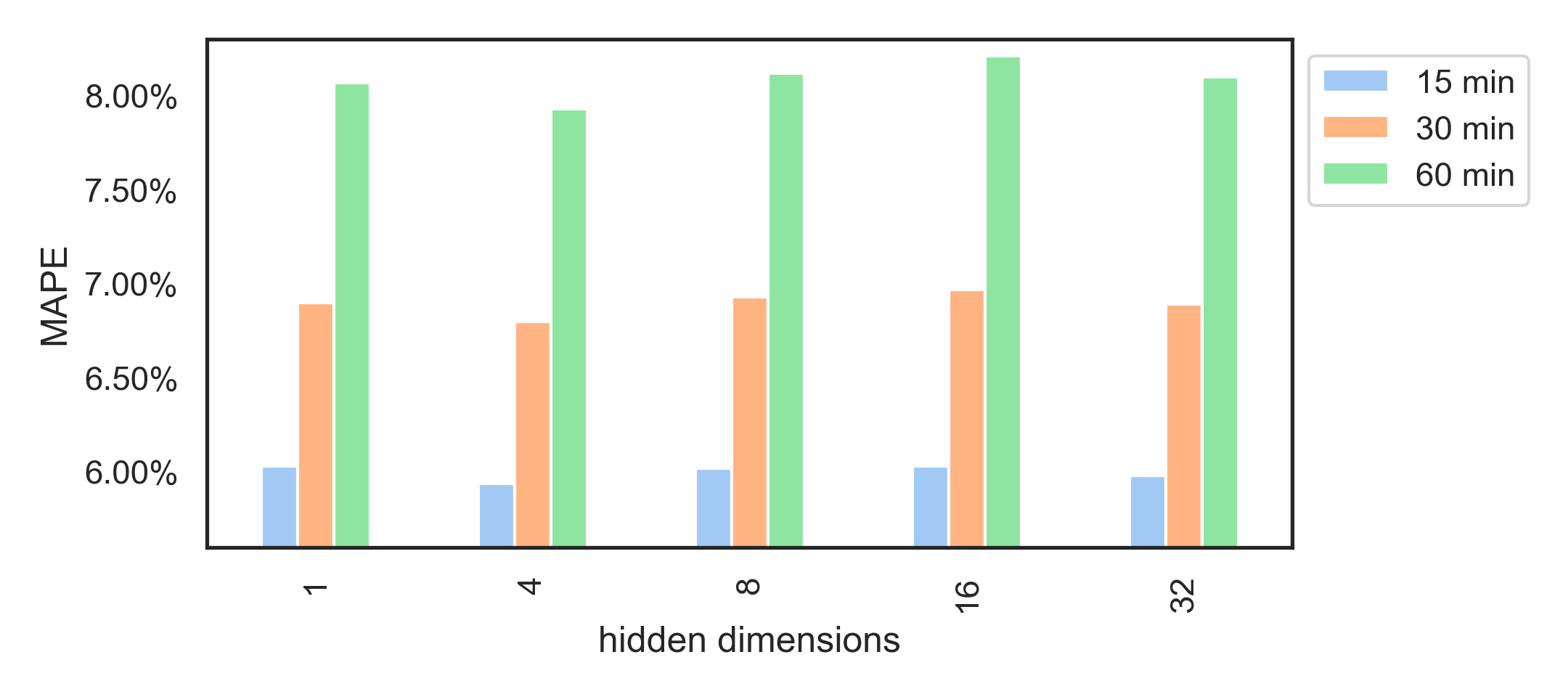}
}
\caption{Loss values of the Transformer model on the validation set during the pretraining stage. (a) Change in validation losses as iteration continues, and (b) the lowest validation loss values during iterations with different dimensions of subseries representations.}
\label{fig:hidden-dim-performance}
\end{figure}

(2)	Dimension of the hidden layers in the long-term trend module and periodicity module. We found the optimal hidden layer dimension from [1, 4, 8, 16, 32] and conducted experiments on the METR-LA dataset. From Figure \ref{fig:hidden-dim-performance}, it can be observed that when the dimension is set to 4, the value of each metric reaches the optimum, while a larger dimension causes a performance drop. A small dimension is not sufficient for the model to fully capture the trend and periodic features in the data, while a large dimension may cause the model to overfit the training set and reduce the generalizability, resulting in poor model performance on the test set. Therefore, we set the hidden layer dimension of the long-term trend module and the periodicity module to 4 to obtain the best performance.

\subsubsection{Visualization}
\begin{figure}[htbp]
\centering  
\subfigure{
\includegraphics[width=5 cm]{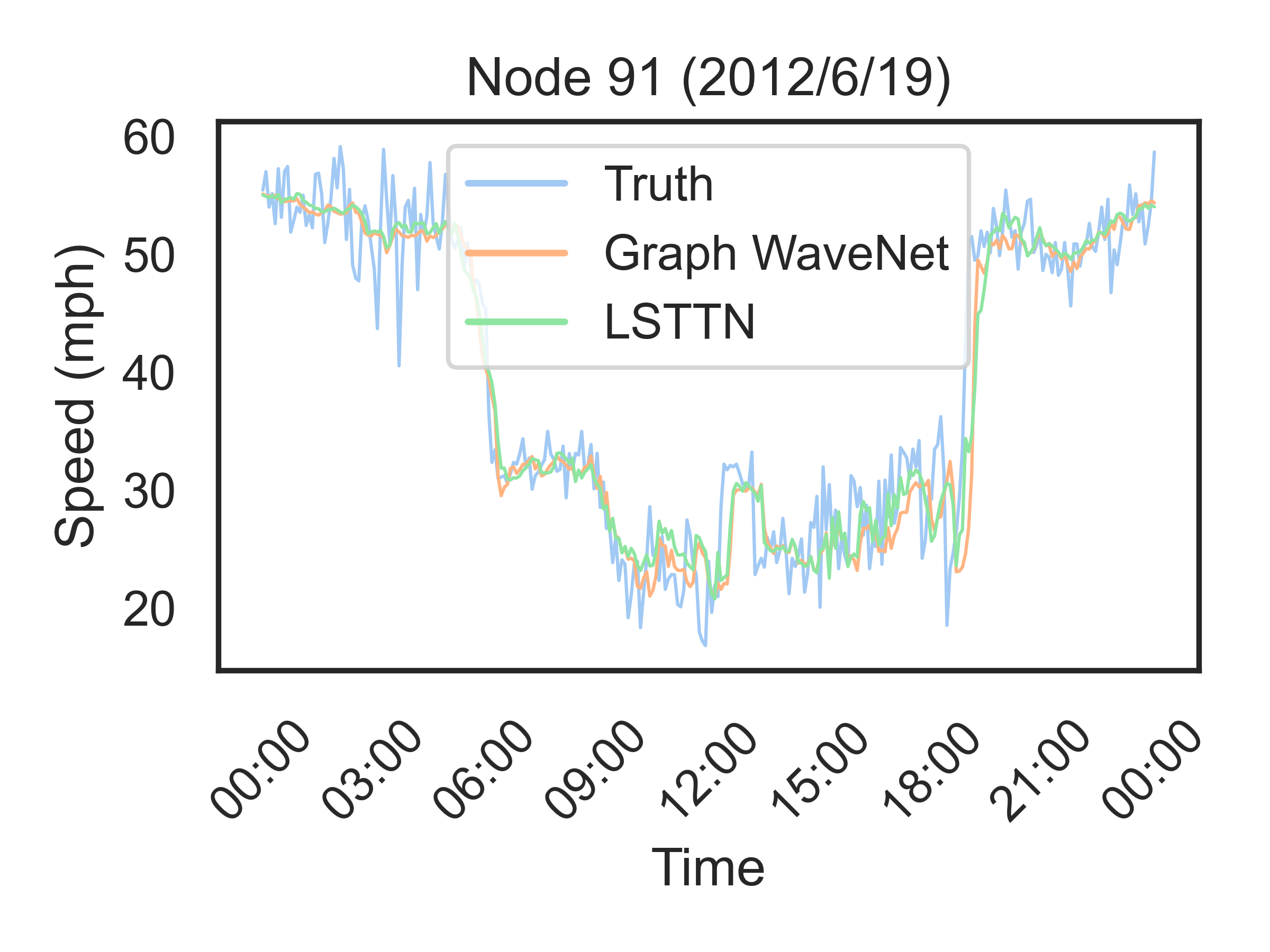}
}
\subfigure{
\includegraphics[width=8.5cm]{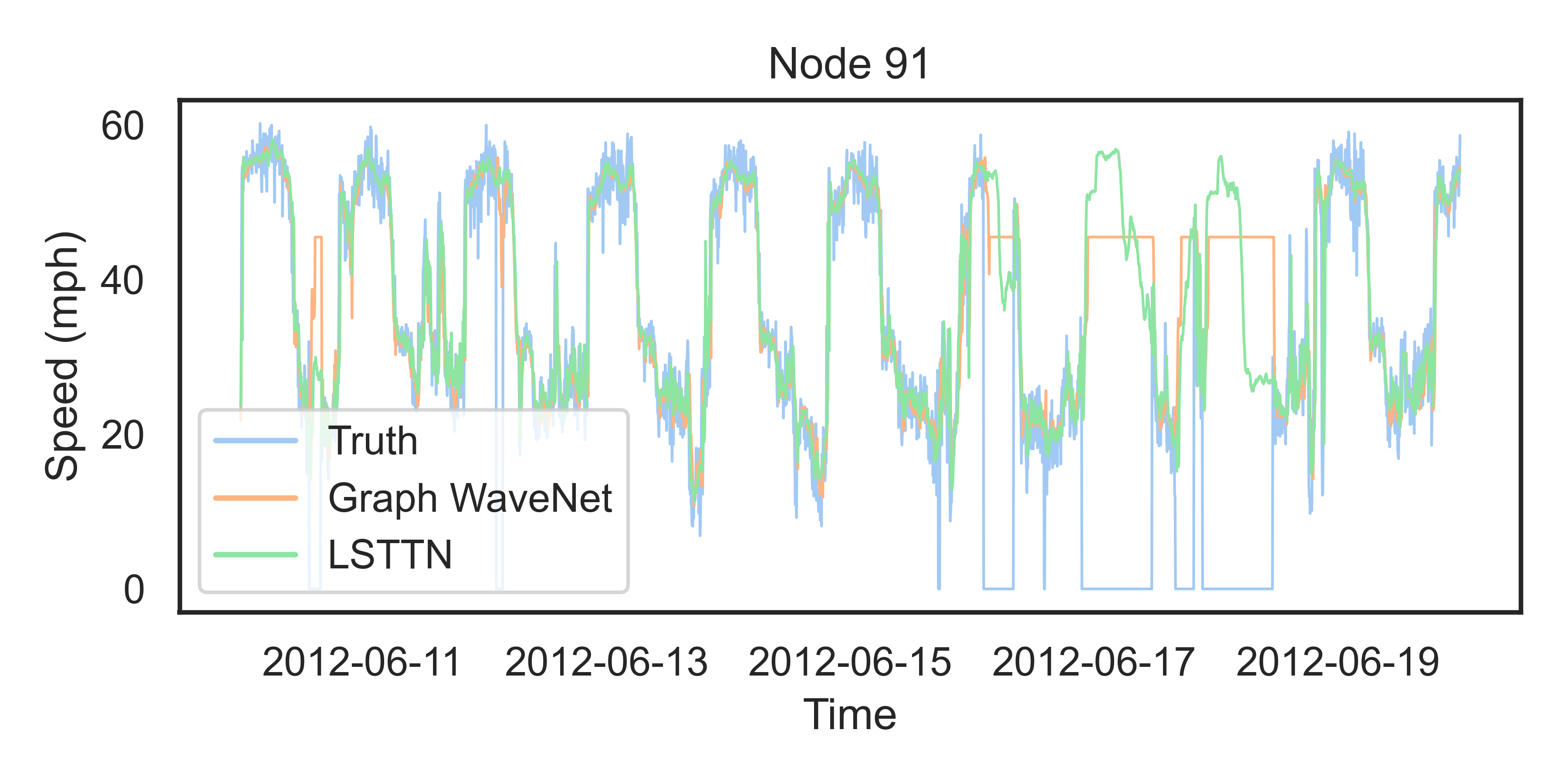}
}
\subfigure{
\includegraphics[width=8.5cm]{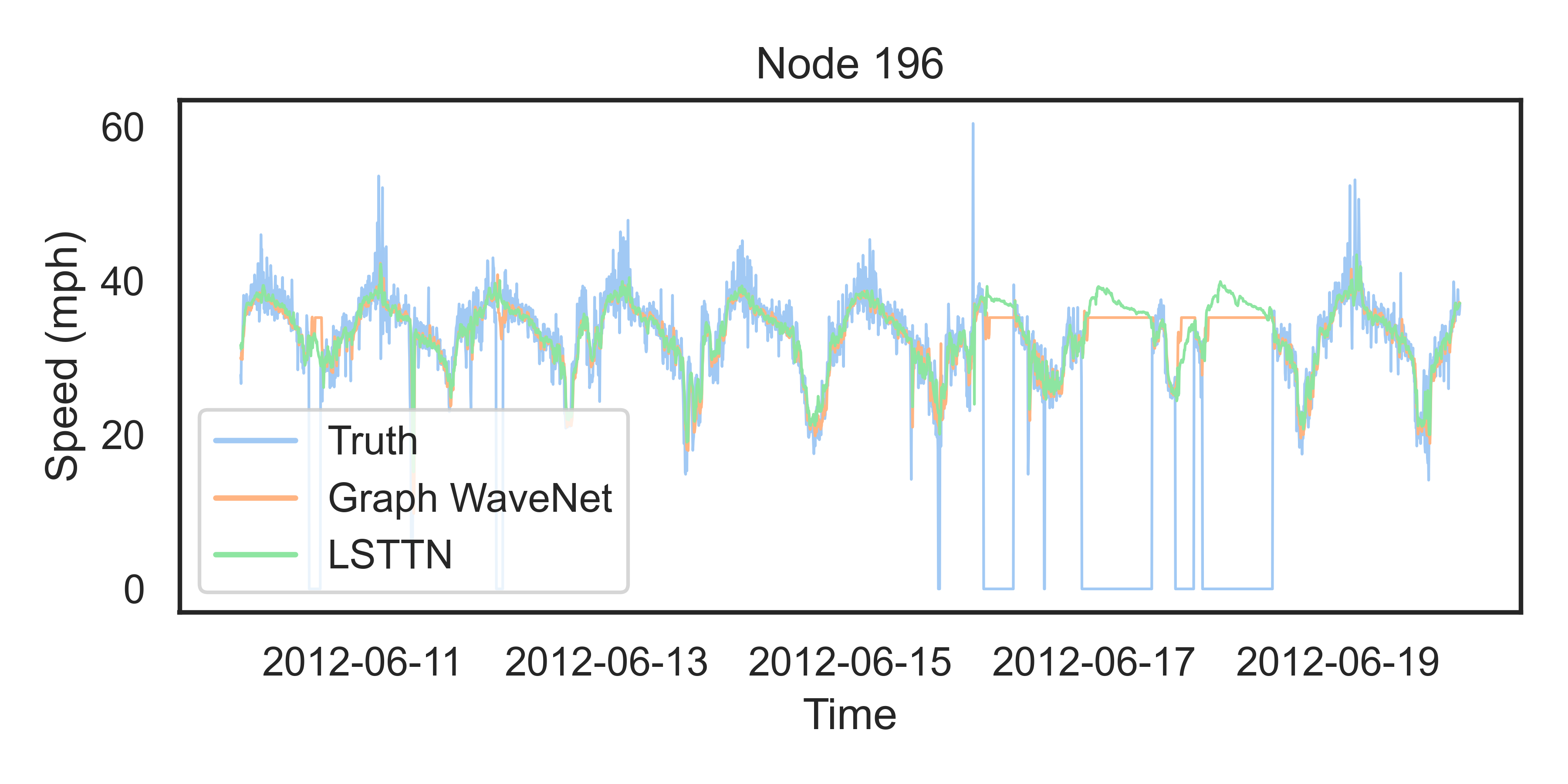}
}
\caption{Visualization of 15-minute ahead predictions on a snapshot of the METR-LA test set}
\label{fig:visualizations-15min}
\end{figure}

\begin{figure}[htbp]
\centering  
\subfigure{
\includegraphics[width=5 cm]{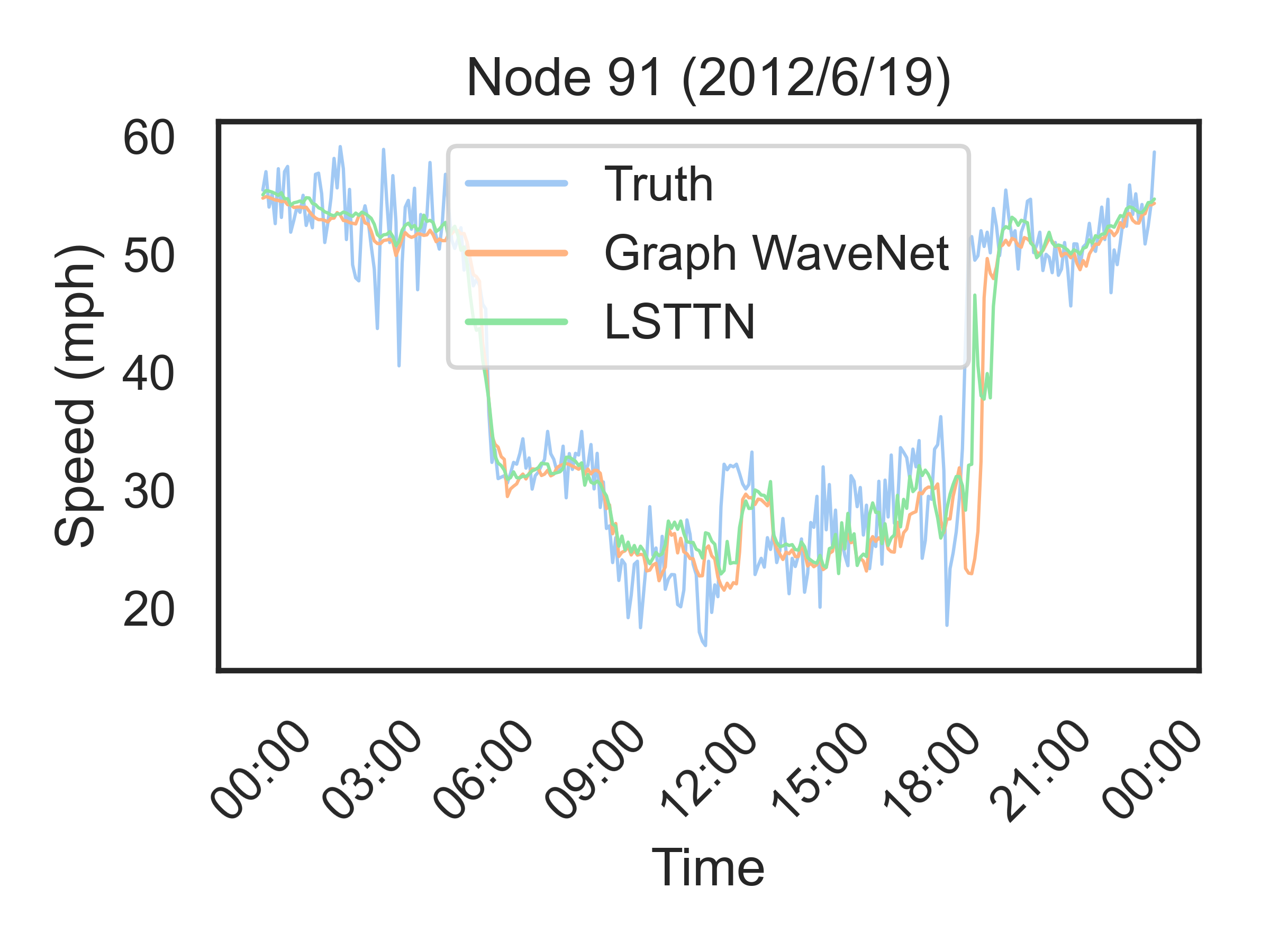}
}
\subfigure{
\includegraphics[width=8.5cm]{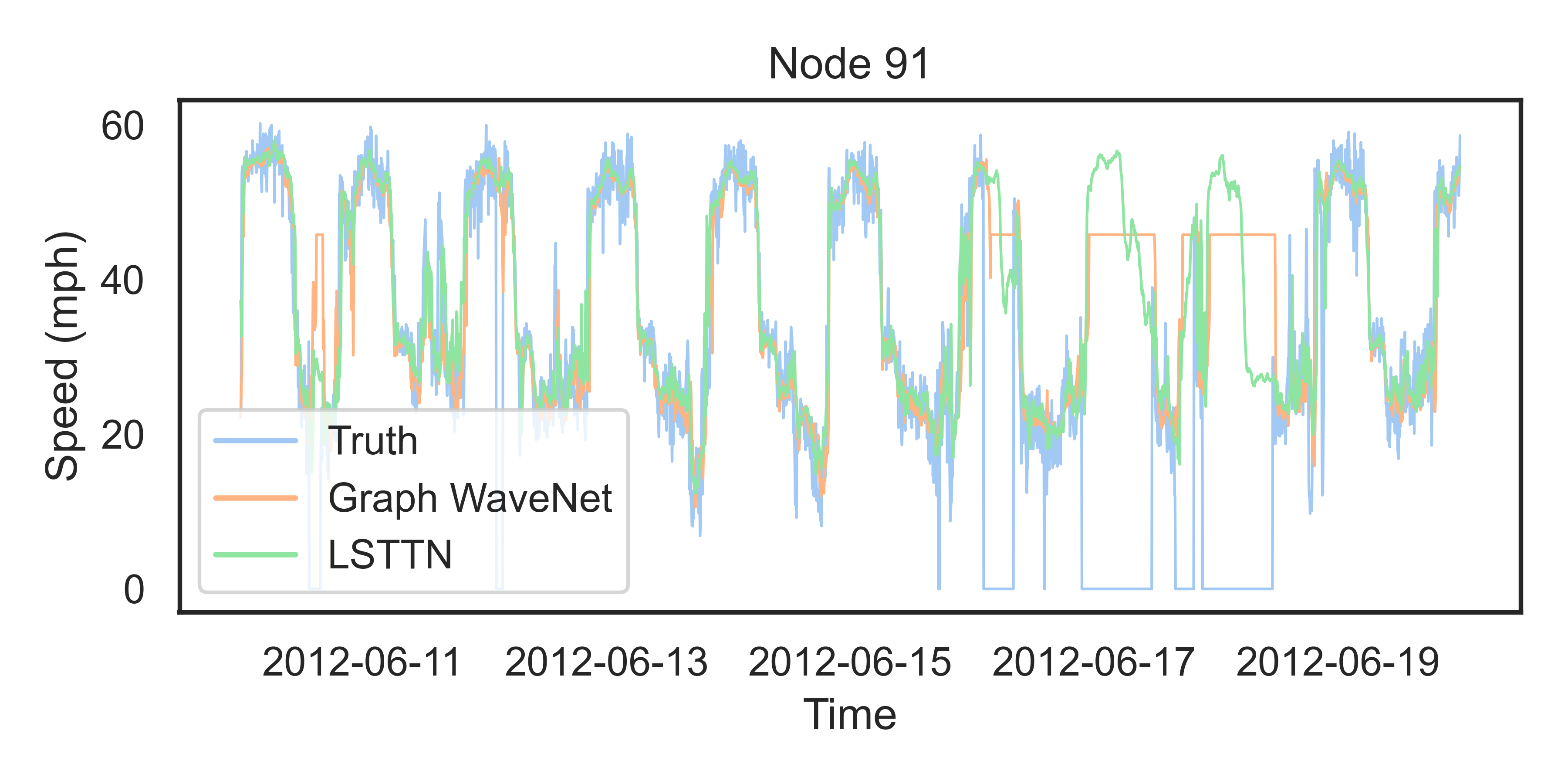}
}
\subfigure{
\includegraphics[width=8.5cm]{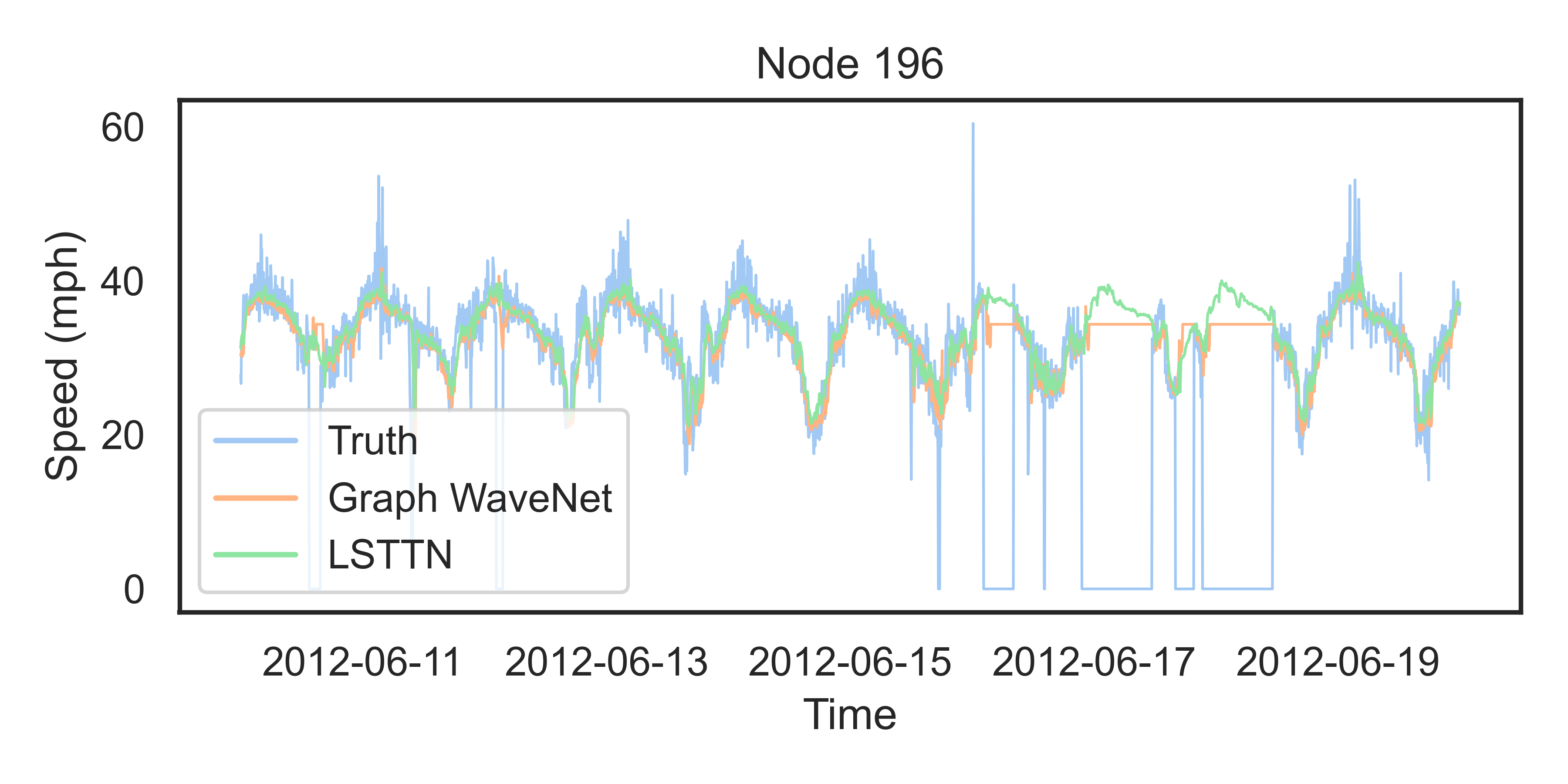}
}
\caption{Visualization of 30-minute ahead predictions on a snapshot of the METR-LA test set}
\label{fig:visualizations-30min}
\end{figure}

\begin{figure}[htbp]
\centering  
\subfigure{
\includegraphics[width=5 cm]{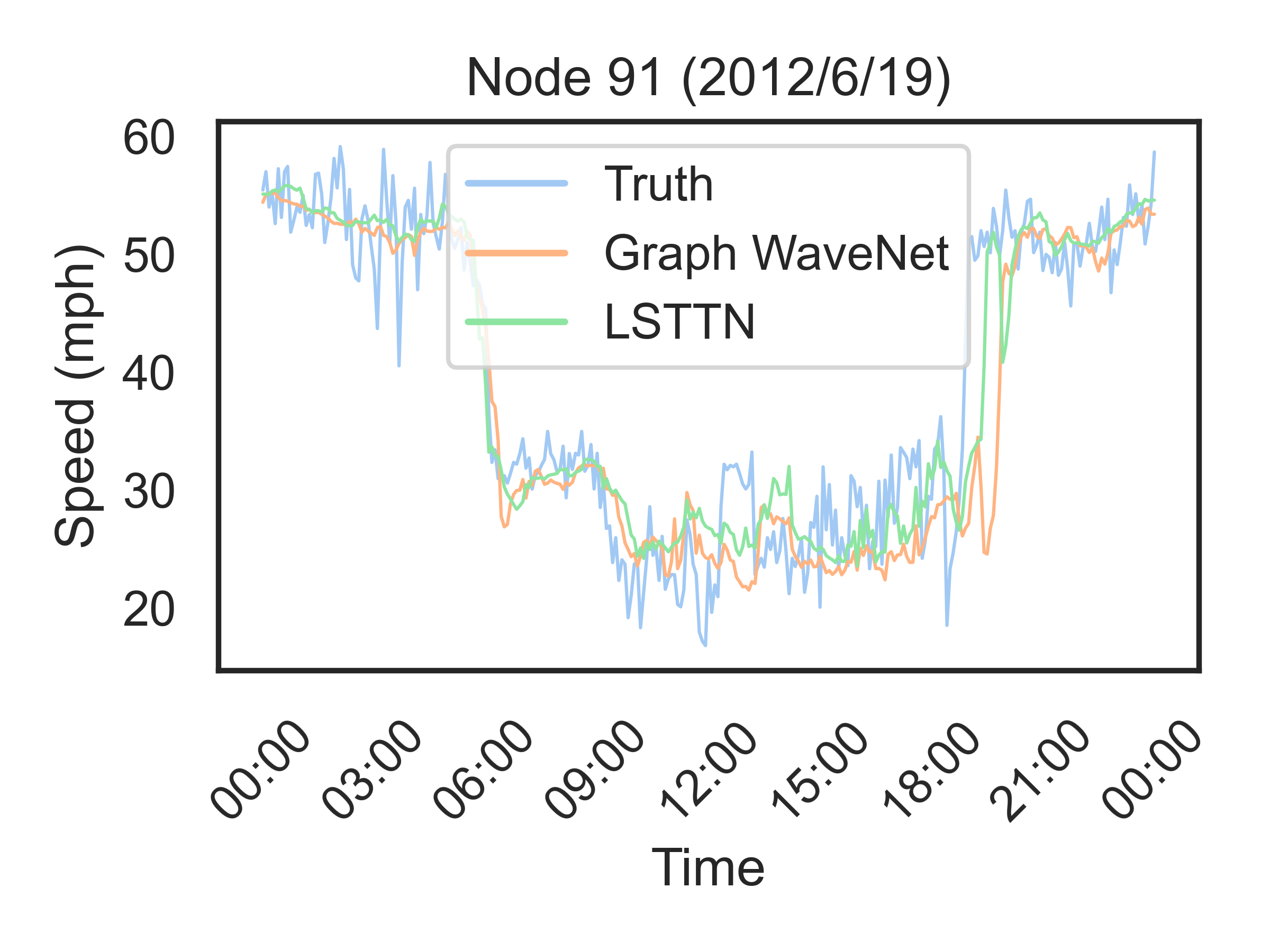}
}
\subfigure{
\includegraphics[width=8.5cm]{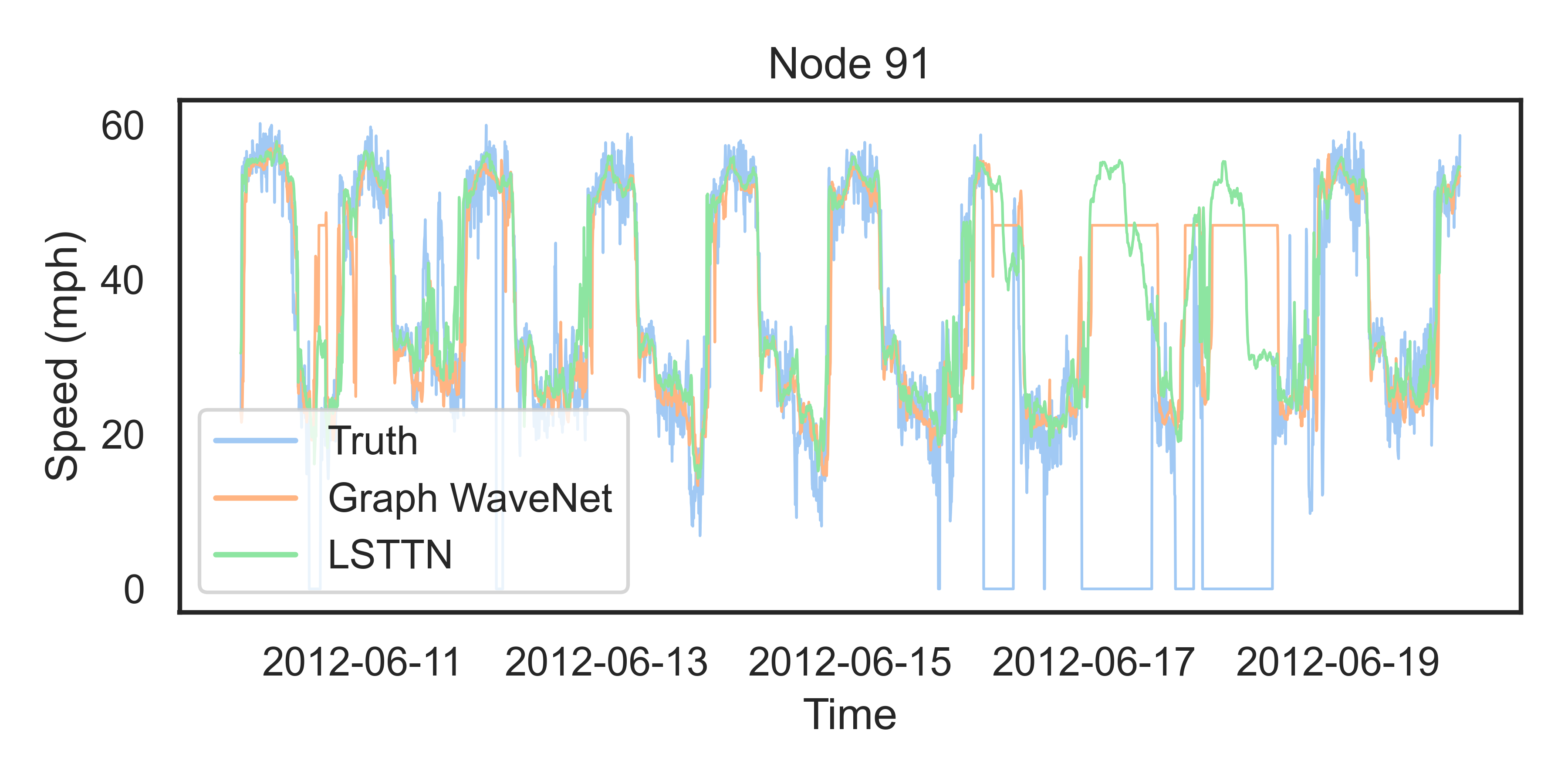}
}
\subfigure{
\includegraphics[width=8.5cm]{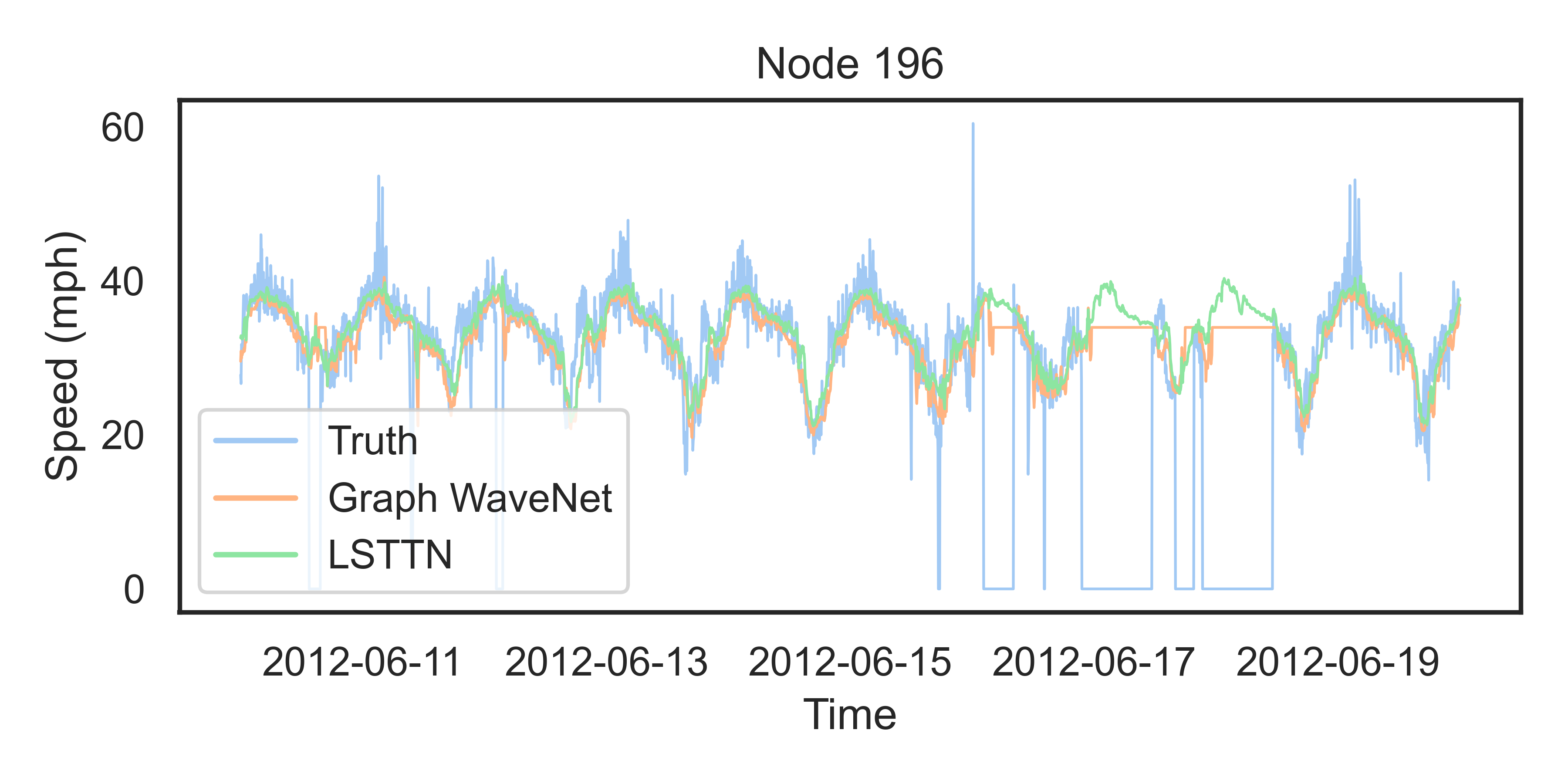}
}
\caption{Visualization of 60-minute ahead predictions on a snapshot of the METR-LA test set}
\label{fig:visualizations-60min}
\end{figure}
To demonstrate the performance and advantages of the LSTTN model more clearly, we visualize the 15-minute ahead, 30-minute ahead and 60-minute ahead prediction results of the LSTTN model and Graph WaveNet on the METR-LA dataset. Figure $8\sim10$ show the prediction results for sensor 773023 and sensor 771673 (corresponding to node 91 and node 196, respectively) from June 10 (Sunday) to June 19, 2012 (Tuesday). The zero values in the ground truth indicate missing data. It can be observed that the LSTTN model is able to provide accurate prediction results for all prediction horizons. In addition, the LSTTN model has two major advantages:

(1) LSTTN can sense sudden changes in traffic flow more quickly.

To further compare the performance from a smaller scale, we visualize the prediction of sensor 773023 (node 91) on June 19. It can be found that the accuracy of both the LSTTN model and Graph WaveNet decreases as we increase the horizon, and the accuracy drops more significantly in the case of large fluctuations. However, the deviation in the results provided by the LSTTN model is smaller in comparison. For example, when the traffic speed rebounded suddenly and significantly at approximately 19:00 as the evening travel peak ended, and such sudden change is neither a random occurrence nor a short-term trend, but a periodic feature , As seen in Figure $8\sim10$, both the LSTTN model and Graph WaveNet exhibited a certain time delay in predicting the sudden change, and the time delay was more significant when the prediction horizon was larger. However, the time delay for the LSTTN model is relatively smaller, and the predicted value is closer to the ground truth. Considering that Graph WaveNet is designed to learn short-term trend only, the performance difference between Graph Wavenet and the LSTTN model may be attributed to that Graph WaveNet mistakenly regards the sudden change in the short-term sequence as a noise instead of a feature, whereas the LSTTN model captures this feature by the periodicity extractor and utilizes it for predictions.

(2) LSTTN can make rough predictions even when short-time historical data are missing.

Missing data often occurs in traffic road network data due to communication failures, sensor malfunctions or system maintenance. For example, the network-wide level traffic data are missing on June 9, 17 and 18, 2012, in the METR-LA dataset. the LSTTN model is able to make rough prediction results based on the temporal information provided by the long historical time series even when a period of data is missing. From the prediction result shown in Figure \ref{fig:visualizations-60min}, we can see that since the input window of the ordinary STGNN (e.g., Graph WaveNet) is short, when the length of missing data is longer than the STGNN input window, the input time series will only contain zeros, and the ordinary STGNNs cannot utilize any valid temporal information and can only output an average value without any fluctuation. However, the LSTTN model can still provide meaningful prediction results in this case. For instance, the LSTTN model successfully predicted the spiky temporal pattern of node 91; for node 196, whose speed changes more smoothly, the LSTTN model can predict the trend of rapid increase and then slow decrease during the missing data period. This result indicates that the LSTTN model can effectively utilize long-term trends and the periodic information provided by the long historical time series to obtain rough prediction results even when there is a long period of missing data.

\subsubsection{Execution cost analysis}
We analyse the execution costs of the LSTTN model and the three most comparative baseline models on METR-LA using the same hardware.
For a better interpretation of the execution-cost comparison (Table \ref{COST}), a detailed analysis into the traning time of the LSTTN model (Table \ref{time analysis}) and a brief comparison of model design (Table \ref{model design}) are provided.
According to Table \ref{COST}, the LSTTN model requires a larger number of parameters and longer training time than the other models. Basically, the increases in execution cost come from three aspects:
\begin{enumerate}
    \item Employment of Transformer. For the LSTTN model, over 96\% of its training time is spent on the masked sub-series Transformer (Table \ref{time analysis}). Compared with graph convolution network and graph adjacency matrix,  Transformer requires considerably longer training time. That is also reflected in the result that besides the LSTTN model, DSTET which also employs Transformer demands significantly longer time for training than the other two models without Transformer (Table \ref{COST}). As for the gap between the LSTTN model and DSTET, it is mainly a consequence of difference in the length of input sequence. As noted in Table 1, the time complexity of Transformer is proportional to the length of input sequence ($L_s$). Since DSTET utilizes Transformer to learn the short-term features, while the Transformer employed in the LSTTN model is for better learning of long-term features, the input sequence of the latter Transformer (4032 time steps) is hundreds of times longer than that of the former Transformer (12 time steps). That makes training the latter Transformer cost much more time than training the latter one.
    \item Additional learning of long-term features. As Table \ref{model design} indicates, among the four models, the LSTTN model is the only one learning both short-term features (short-term trend) and long-term features (long-term trend and periodicity). Two more feature-capture modules, namely long-term extractor and periodicity extractor, are deployed in the LSTTN model, compared with the other models. Training the two modules require additional time and parameters.
    \item Additional training of feature fusion. As noted earlier, the LSTTN models learns three types of features. To better utilize the learned features for better forecasting, a feature-fusion module is also deployed in the LSTTN model. This deployment brings not only additional parameters but also additional stage of training to the model.
\end{enumerate}

\begin{table*}[htbp]
\caption{\centering {Execution costs of models on METR-LA under a NVIDIA RTX A6000 GPU}}
\label{COST}
\centering
\begin{threeparttable}
\begin{tabular}{cccccc}
\hline
                & Model   & Time complexity    & Most time-consuming component & Training time (s/epoch)    & Params (M)  \\ \hline
                & Graph WaveNet   & $O({n_o}^2 + {n_e})$   & {\begin{tabular}[c]{@{}c@{}}Graph adjacency matrix,\\ Graph convolution network\end{tabular}} & 19.6   & 0.99  \\
                & MTGNN   & $O({n_o}^2 + {n_e})$   & {\begin{tabular}[c]{@{}c@{}}Graph adjacency matrix,\\ Graph convolution network\end{tabular}} & 22.3   & 1.61   \\
                & DSTET   & $O({L_s}^2{e_d})$   & Transformer & 48.2   & 0.72   \\
                & LSTTN (ours)   & $O({L_s}^2{e_d})$   & Transformer & 611.6   & 2.19  \\ \hline
\end{tabular}
\begin{tablenotes}
\footnotesize
\item[*] For each model, its estimated trainig time is the sum of time used to train its each module.
\item[**] The time complexities of graph adjacency matrix and graph convolution network are $O({n_o}^2)$ and $O({n_e}^2)$, $n_o$ is the number of nodes, $n_e$ is the number of edges, $L_s$ denotes the length of input sequence, $e_d$ denotes the dimension of embedding.
\item[***] The time complexity of Transformer is $O({L_s}^2{e_d})$, $L_s$ denotes the length of input sequence, $e_d$ denotes the dimension of embedding.
\end{tablenotes}
\end{threeparttable}
\end{table*}

\begin{table*}[htbp]
\caption{\centering {Training time analysis of LSTNN model on METR-LA under a NVIDIA RTX A6000 GPU}}
\label{time analysis}
\centering
\begin{tabular}{ccc}
\hline
                & Module   & Percentage of the overall training time\\ \hline
                & Masked Sub-series Transformer   & 96.50\%  \\
                & Long-term Trend Extractor   & 1.20\%   \\
                & Periodicity Extractor   & 0.21\%   \\
                & Short-term Trend Extractor   & 2.08\%   \\
                & Feature Fusion   & 0.01\%  \\ \hline
\end{tabular}
\end{table*}

\begin{table*}[htbp]
\caption{\centering {A brief comparison of model design}}
\label{model design}
\resizebox{\linewidth}{!}{
\begin{tabular}{cccccccc}
\hline
\multicolumn{2}{c}{\multirow{2}{*}{\textbf{}}}                                 & \multicolumn{3}{c}{Components}                         & \multicolumn{3}{c}{Learned features}                          \\
\multicolumn{2}{c}{}                                                           & Graph adjacency matrix          & Graph convolution network          & Transformer          & Short-term trend          & Periodicity          & Long-term trend          \\ \hline
&Graph WaveNet                 & \Checkmark   & \Checkmark    & \XSolidBrush   & \Checkmark & \XSolidBrush & \XSolidBrush \\
&MTGNN                 & \Checkmark   & \Checkmark    & \XSolidBrush   & \Checkmark & \XSolidBrush & \XSolidBrush \\
&DSTET                 & \Checkmark   & \XSolidBrush    & \Checkmark   & \Checkmark & \XSolidBrush & \XSolidBrush \\
&LSTTN (ours)                 & \Checkmark   & \Checkmark    & \Checkmark   & \Checkmark & \Checkmark & \Checkmark \\ \hline
\end{tabular}
}
\end{table*}

\section{Conclusion and Future Work}
\label{sec:conclusion}
In consideration of the insufficiency of short-term features for traffic forecasting, this paper proposes a novel framework named LSTTN that integrates both the long-term features and short-term features present in historical traffic data to obtain more accurate prediction results. The proposed LSTTN is pluggable for its compatibility with all STGNN models, indicating the feasibility of packing LSTTN as a plugin to make it easier to deploy LSTTN to application platforms.

Under the proposed framework, a specific model design is also provided. Apart from extracting short-term features using STGNN, the LSTTN model also learns two additional long-term features, namely long-term trend and periodicity. And for valid extraction of long-term features, the LSTTN model first learns sub-series temporal representations via a masked subseries Transformer, then based on the learned representations, it extracts long-term trend and periodicity utilizing stacked 1D dilated convolution and spatial-based graph convolution, respectively. Besides, a specific strategy for fusing long-term and short term features is also employed for better prediction. Experiments on real-world datasets with different prediction horizons show that the LSTTN model achieves better performance than other baseline models, which implies the effectiveness of integrating long-term and short term features to improve the prediction accuracy. The conducted ablation experiments verify the validity of each learning module of the LSTTN model, and the visualization of the prediction results show that the LSTTN model is capable of perceiving upcoming abrupt changes faster and provide prediction results regardless of missing data. Those results also corroborate the role of long-term features, namely long-term trend and periodicity can play in traffic forecasting.

Although LSTTN can effectively utilize the temporal information in long historical series and make more accurate predictions, there are still some shortcomings that deserve further research:
\begin{enumerate}
    \item The evaluation of LSTTN mainly concerns the prediction accuracy, whereas high prediction accuracy alone may not guarantee its practicability. For example, as stated earlier, compared with mainstream STGNNs learning short-term features only such as Graph WaveNet, the LSTTN model requires longer training time and more parameters, which will influence the adoption of LSTTN in real-world scenarios. Besides solution to the additional training cost, issues like user-friendliness should also be well considered in the future.
    \item The design of the subseries temporal representation learner, the long-term trend extractor, the periodicity extractor and the short-term trend extractor in the LSTTN framework and the strategy used to fuse long- and short-term features can all be regarded as feasible concrete implementations, and other forms of module design and fusion strategies can be further explored. For example, future exploration can be done on learning dynamic spatial dependencies for the extraction of periodic features.
    \item Although ablation studies and visualization analysis are performed to improve the transparency and interpretability of the performance of the LSTTN model, there is still certain room for improvement.
\end{enumerate}

\section*{CRediT authorship contribution statement}
\textbf{Qinyao Luo}: Conceptualization, Methodology, Writing – original draft, Writing - Review \& Editing.
\textbf{Silu He}: Data curation, Writing – original draft, Writing - Review \& Editing. 
\textbf{Xing Han}: Software, Validation. 
\textbf{Xuhan Wang}: Data curation, Writing – original draft. 
\textbf{Haifeng Li}: Conceptualization, Methodology, Writing – original draft, Writing - Review \& Editing, Supervision, Funding acquisition.

\section*{Declaration of competing interest}
The authors declare that they have no known competing financial interests or personal relationships that could have appeared to influence the work reported in this paper.

\section*{Data availability}
Data is available at https://github.com/GeoX-Lab/LSTTN.

\section*{Acknowledgement}
This work was supported by the National Natural Science
Foundation of China [grant numbers 41871364, 4227148 1], and Hunan Provincial Natural Science Foundation of China [grant number 2022JJ30698].





\bibliographystyle{elsarticle-num} 
\bibliography{references}

\begin{thebibliography}{10}
\expandafter\ifx\csname url\endcsname\relax
  \def\url#1{\texttt{#1}}\fi
\expandafter\ifx\csname urlprefix\endcsname\relax\def\urlprefix{URL }\fi
\expandafter\ifx\csname href\endcsname\relax
  \def\href#1#2{#2} \def\path#1{#1}\fi

\bibitem{yu2004switching}
G.~Yu, C.~Zhang, Switching arima model based forecasting for traffic flow, in: 2004 IEEE International Conference on Acoustics, Speech, and Signal Processing, Vol.~2, IEEE, 2004, pp. ii--429.

\bibitem{chandra2009predictions}
S.~R. Chandra, H.~Al-Deek, Predictions of freeway traffic speeds and volumes using vector autoregressive models, Journal of Intelligent Transportation Systems 13~(2) (2009) 53--72.

\bibitem{zhang2017deep}
J.~Zhang, Y.~Zheng, D.~Qi, Deep spatio-temporal residual networks for citywide crowd flows prediction, in: Proceedings of the AAAI conference on artificial intelligence, Vol.~31, 2017.

\bibitem{shi2015convolutional}
X.~Shi, Z.~Chen, H.~Wang, D.-Y. Yeung, W.-K. Wong, W.-c. Woo, Convolutional lstm network: A machine learning approach for precipitation nowcasting, Advances in neural information processing systems 28 (2015).

\bibitem{li2017diffusion}
Y.~Li, R.~Yu, C.~Shahabi, Y.~Liu, Diffusion convolutional recurrent neural network: Data-driven traffic forecasting, arXiv preprint arXiv:1707.01926 (2017).

\bibitem{yu2017spatio}
B.~Yu, H.~Yin, Z.~Zhu, Spatio-temporal graph convolutional networks: A deep learning framework for traffic forecasting, arXiv preprint arXiv:1709.04875 (2017).

\bibitem{smith1997traffic}
B.~L. Smith, M.~J. Demetsky, Traffic flow forecasting: comparison of modeling approaches, Journal of transportation engineering 123~(4) (1997) 261--266.

\bibitem{gao2013application}
J.~W. Gao, Z.~W. Leng, B.~Zhang, X.~Liu, G.~Q. Cai, The application of adaptive kalman filter in traffic flow forecasting, in: Advanced Materials Research, Vol. 680, Trans Tech Publ, 2013, pp. 495--500.

\bibitem{jiang2022graph}
W.~Jiang, J.~Luo, Graph neural network for traffic forecasting: A survey, Expert Systems with Applications (2022) 117921.

\bibitem{ma2017learning}
X.~Ma, Z.~Dai, Z.~He, J.~Ma, Y.~Wang, Y.~Wang, Learning traffic as images: A deep convolutional neural network for large-scale transportation network speed prediction, Sensors 17~(4) (2017) 818.

\bibitem{liu2017short}
Y.~Liu, H.~Zheng, X.~Feng, Z.~Chen, Short-term traffic flow prediction with conv-lstm, in: 2017 9th International Conference on Wireless Communications and Signal Processing (WCSP), IEEE, 2017, pp. 1--6.

\bibitem{kipf2016variational}
T.~N. Kipf, M.~Welling, Variational graph auto-encoders, arXiv preprint arXiv:1611.07308 (2016).

\bibitem{zhang2018end}
M.~Zhang, Z.~Cui, M.~Neumann, Y.~Chen, An end-to-end deep learning architecture for graph classification, in: Proceedings of the AAAI conference on artificial intelligence, Vol.~32, 2018.

\bibitem{li2022curvature}
H.~Li, J.~Cao, J.~Zhu, Y.~Liu, Q.~Zhu, G.~Wu, Curvature graph neural network, Information Sciences 592 (2022) 50--66.

\bibitem{wu2020comprehensive}
Z.~Wu, S.~Pan, F.~Chen, G.~Long, C.~Zhang, S.~Y. Philip, A comprehensive survey on graph neural networks, IEEE transactions on neural networks and learning systems 32~(1) (2020) 4--24.

\bibitem{bruna2013spectral}
J.~Bruna, W.~Zaremba, A.~Szlam, Y.~LeCun, Spectral networks and locally connected networks on graphs, arXiv preprint arXiv:1312.6203 (2013).

\bibitem{defferrard2016convolutional}
M.~Defferrard, X.~Bresson, P.~Vandergheynst, Convolutional neural networks on graphs with fast localized spectral filtering, Advances in neural information processing systems 29 (2016).

\bibitem{kipf2016semi}
T.~N. Kipf, M.~Welling, Semi-supervised classification with graph convolutional networks, arXiv preprint arXiv:1609.02907 (2016).

\bibitem{gilmer2017neural}
J.~Gilmer, S.~S. Schoenholz, P.~F. Riley, O.~Vinyals, G.~E. Dahl, Neural message passing for quantum chemistry, in: International conference on machine learning, PMLR, 2017, pp. 1263--1272.

\bibitem{velivckovic2017graph}
P.~Veli{\v{c}}kovi{\'c}, G.~Cucurull, A.~Casanova, A.~Romero, P.~Lio, Y.~Bengio, Graph attention networks, arXiv preprint arXiv:1710.10903 (2017).

\bibitem{bai2021a3t}
J.~Bai, J.~Zhu, Y.~Song, L.~Zhao, Z.~Hou, R.~Du, H.~Li, A3t-gcn: Attention temporal graph convolutional network for traffic forecasting, ISPRS International Journal of Geo-Information 10~(7) (2021) 485.

\bibitem{he2022stgc}
S.~He, Q.~Luo, R.~Du, L.~Zhao, H.~Li, Stgc-gnns: A gnn-based traffic prediction framework with a spatial-temporal granger causality graph, arXiv preprint arXiv:2210.16789 (2022).

\bibitem{cai2020traffic}
L.~Cai, K.~Janowicz, G.~Mai, B.~Yan, R.~Zhu, Traffic transformer: Capturing the continuity and periodicity of time series for traffic forecasting, Transactions in GIS 24~(3) (2020) 736--755.

\bibitem{park2020st}
C.~Park, C.~Lee, H.~Bahng, Y.~Tae, S.~Jin, K.~Kim, S.~Ko, J.~Choo, St-grat: A novel spatio-temporal graph attention networks for accurately forecasting dynamically changing road speed, in: Proceedings of the 29th ACM international conference on information \& knowledge management, 2020, pp. 1215--1224.

\bibitem{roy2021unified}
A.~Roy, K.~K. Roy, A.~A. Ali, M.~A. Amin, A.~M. Rahman, Unified spatio-temporal modeling for traffic forecasting using graph neural network, in: 2021 International Joint Conference on Neural Networks (IJCNN), IEEE, 2021, pp. 1--8.

\bibitem{zhao2019t}
L.~Zhao, Y.~Song, C.~Zhang, Y.~Liu, P.~Wang, T.~Lin, M.~Deng, H.~Li, T-gcn: A temporal graph convolutional network for traffic prediction, IEEE transactions on intelligent transportation systems 21~(9) (2019) 3848--3858.

\bibitem{zhu2021ast}
J.~Zhu, Q.~Wang, C.~Tao, H.~Deng, L.~Zhao, H.~Li, Ast-gcn: Attribute-augmented spatiotemporal graph convolutional network for traffic forecasting, IEEE Access 9 (2021) 35973--35983.

\bibitem{zhu2022kst}
J.~Zhu, X.~Han, H.~Deng, C.~Tao, L.~Zhao, P.~Wang, T.~Lin, H.~Li, Kst-gcn: A knowledge-driven spatial-temporal graph convolutional network for traffic forecasting, IEEE Transactions on Intelligent Transportation Systems 23~(9) (2022) 15055--15065.

\bibitem{wu2019graph}
Z.~Wu, S.~Pan, G.~Long, J.~Jiang, C.~Zhang, Graph wavenet for deep spatial-temporal graph modeling, arXiv preprint arXiv:1906.00121 (2019).

\bibitem{guo2019attention}
S.~Guo, Y.~Lin, N.~Feng, C.~Song, H.~Wan, Attention based spatial-temporal graph convolutional networks for traffic flow forecasting, in: Proceedings of the AAAI conference on artificial intelligence, Vol.~33, 2019, pp. 922--929.

\bibitem{vaswani2017attention}
A.~Vaswani, N.~Shazeer, N.~Parmar, J.~Uszkoreit, L.~Jones, A.~N. Gomez, {\L}.~Kaiser, I.~Polosukhin, Attention is all you need, Advances in neural information processing systems 30 (2017).

\bibitem{xu2020spatial}
M.~Xu, W.~Dai, C.~Liu, X.~Gao, W.~Lin, G.-J. Qi, H.~Xiong, Spatial-temporal transformer networks for traffic flow forecasting, arXiv preprint arXiv:2001.02908 (2020).

\bibitem{zhou2021informer}
H.~Zhou, S.~Zhang, J.~Peng, S.~Zhang, J.~Li, H.~Xiong, W.~Zhang, Informer: Beyond efficient transformer for long sequence time-series forecasting, in: Proceedings of the AAAI conference on artificial intelligence, Vol.~35, 2021, pp. 11106--11115.

\bibitem{zeng2022transformers}
A.~Zeng, M.~Chen, L.~Zhang, Q.~Xu, Are transformers effective for time series forecasting?, arXiv preprint arXiv:2205.13504 (2022).

\bibitem{he2022masked}
K.~He, X.~Chen, S.~Xie, Y.~Li, P.~Doll{\'a}r, R.~Girshick, Masked autoencoders are scalable vision learners, in: Proceedings of the IEEE/CVF Conference on Computer Vision and Pattern Recognition, 2022, pp. 16000--16009.

\bibitem{chen2021self}
D.~Chen, Y.~Chen, Y.~Li, F.~Mao, Y.~He, H.~Xue, Self-supervised learning for few-shot image classification, in: ICASSP 2021-2021 IEEE International Conference on Acoustics, Speech and Signal Processing (ICASSP), IEEE, 2021, pp. 1745--1749.

\bibitem{devlin2018bert}
J.~Devlin, M.-W. Chang, K.~Lee, K.~Toutanova, Bert: Pre-training of deep bidirectional transformers for language understanding, arXiv preprint arXiv:1810.04805 (2018).

\bibitem{radford2019language}
A.~Radford, J.~Wu, R.~Child, D.~Luan, D.~Amodei, I.~Sutskever, et~al., Language models are unsupervised multitask learners, OpenAI blog 1~(8) (2019) 9.

\bibitem{li2023augmentation}
H.~Li, J.~Cao, J.~Zhu, Q.~Luo, S.~He, X.~Wang, Augmentation-free graph contrastive learning of invariant-discriminative representations, IEEE Transactions on Neural Networks and Learning Systems (2023).

\bibitem{zhu2022high}
J.~Zhu, B.~Li, Z.~Zhang, L.~Zhao, H.~Li, High-order topology-enhanced graph convolutional networks for dynamic graphs, Symmetry 14~(10) (2022) 2218.

\bibitem{zhu2022alleviating}
J.~Zhu, M.~Hong, R.~Du, H.~Li, Alleviating neighbor bias: augmenting graph self-supervise learning with structural equivalent positive samples, arXiv preprint arXiv:2212.04365 (2022).

\bibitem{liu2022contrastive}
X.~Liu, Y.~Liang, C.~Huang, Y.~Zheng, B.~Hooi, R.~Zimmermann, When do contrastive learning signals help spatio-temporal graph forecasting?, in: Proceedings of the 30th International Conference on Advances in Geographic Information Systems, 2022, pp. 1--12.

\bibitem{ji2022spatio}
J.~Ji, J.~Wang, C.~Huang, J.~Wu, B.~Xu, Z.~Wu, J.~Zhang, Y.~Zheng, Spatio-temporal self-supervised learning for traffic flow prediction, arXiv preprint arXiv:2212.04475 (2022).

\bibitem{shao2022pre}
Z.~Shao, Z.~Zhang, F.~Wang, Y.~Xu, Pre-training enhanced spatial-temporal graph neural network for multivariate time series forecasting, in: Proceedings of the 28th ACM SIGKDD Conference on Knowledge Discovery and Data Mining, 2022, pp. 1567--1577.

\bibitem{li2023ti}
Z.~Li, Z.~Rao, L.~Pan, P.~Wang, Z.~Xu, Ti-mae: Self-supervised masked time series autoencoders, arXiv preprint arXiv:2301.08871 (2023).

\bibitem{nie2022time}
Y.~Nie, N.~H. Nguyen, P.~Sinthong, J.~Kalagnanam, A time series is worth 64 words: Long-term forecasting with transformers, arXiv preprint arXiv:2211.14730 (2022).

\bibitem{yu2015multi}
F.~Yu, V.~Koltun, Multi-scale context aggregation by dilated convolutions, arXiv preprint arXiv:1511.07122 (2015).

\bibitem{zheng2020gman}
C.~Zheng, X.~Fan, C.~Wang, J.~Qi, Gman: A graph multi-attention network for traffic prediction, in: Proceedings of the AAAI conference on artificial intelligence, Vol.~34, 2020, pp. 1234--1241.

\bibitem{wu2020connecting}
Z.~Wu, S.~Pan, G.~Long, J.~Jiang, X.~Chang, C.~Zhang, Connecting the dots: Multivariate time series forecasting with graph neural networks, in: Proceedings of the 26th ACM SIGKDD international conference on knowledge discovery \& data mining, 2020, pp. 753--763.

\bibitem{sun2022dual}
Y.~Sun, X.~Jiang, Y.~Hu, F.~Duan, K.~Guo, B.~Wang, J.~Gao, B.~Yin, Dual dynamic spatial-temporal graph convolution network for traffic prediction, IEEE Transactions on Intelligent Transportation Systems 23~(12) (2022) 23680--23693.

\bibitem{shao2022spatial}
Z.~Shao, Z.~Zhang, F.~Wang, W.~Wei, Y.~Xu, Spatial-temporal identity: A simple yet effective baseline for multivariate time series forecasting, in: Proceedings of the 31st ACM International Conference on Information \& Knowledge Management, 2022, pp. 4454--4458.

\bibitem{sun2023transformer}
W.~Sun, R.~Cheng, Y.~Jiao, J.~Gao, Z.~Zheng, N.~Lu, Transformer network with decoupled spatial--temporal embedding for traffic flow forecasting, Applied Intelligence (2023) 1--21.

\bibitem{ouyang2023domain}
X.~Ouyang, Y.~Yang, Y.~Zhang, W.~Zhou, J.~Wan, S.~Du, Domain adversarial graph neural network with cross-city graph structure learning for traffic prediction, Knowledge-Based Systems 278 (2023) 110885.

\bibitem{paszke2019pytorch}
A.~Paszke, S.~Gross, F.~Massa, A.~Lerer, J.~Bradbury, G.~Chanan, T.~Killeen, Z.~Lin, N.~Gimelshein, L.~Antiga, et~al., Pytorch: An imperative style, high-performance deep learning library, Advances in neural information processing systems 32 (2019).

\bibitem{kingma2014adam}
D.~P. Kingma, J.~Ba, Adam: A method for stochastic optimization, arXiv preprint arXiv:1412.6980 (2014).

\bibitem{wang2021libcity}
J.~Wang, J.~Jiang, W.~Jiang, C.~Li, W.~X. Zhao, Libcity: An open library for traffic prediction, in: Proceedings of the 29th International Conference on Advances in Geographic Information Systems, 2021, pp. 145--148.

\end{thebibliography}

\end{document}